\documentclass[times,review,10pt]{elsarticle}
\usepackage{setspace}  % 导入 setspace 宏包
\usepackage{cite}
\usepackage{amsmath,amssymb,amsfonts}
\usepackage{graphicx}
\usepackage{textcomp}
\usepackage{epstopdf}
\usepackage{subfigure}
\usepackage{lineno}
\usepackage{url}
\usepackage{hyperref}
\usepackage[linesnumbered,ruled]{algorithm2e}
\usepackage{xcolor}
\usepackage{colortbl, booktabs} 
\usepackage{framed,multirow}
\usepackage{float}

\journal{Pattern Recognition}

\begin{document}

\begin{frontmatter}
\title{Innovative Tooth Segmentation Using Hierarchical Features and Bidirectional Sequence Modeling} %% Article title

\tnotetext[1]{This work was supported by the Zhejiang Province Natural Science Foundation (No. LZ24F020001), the Opening Foundation of the Tongxiang Institute of General Artificial Intelligence (No. TAGI2-B-2024-0009), and the State Key Laboratory of Advanced Medical Materials and Devices (No. SQ2022SKL01089-2025-14).}
\author[1]{Xinxin Zhao}
\author[2,3]{Jian Jiang}
\author[1,6,9]{Yan Tian\corref{cor1}}
\ead{tianyan@zjgsu.edu.cn}
\author[4]{Liqin Wu}
\author[5]{Zhaocheng Xu}
\author[7]{Teddy Yang}
\author[8]{Yunuo Zou}
\author[1]{Xun Wang}

\affiliation[1]{organization={School of Computer Science and Technology, Zhejiang Gongshang University},
                postcode={310018}, 
                city={Hangzhou},
                country={China}}
\affiliation[2]{organization={School of Computer and Cyber Sciences, Communication University of China},
                postcode={100024}, 
                city={Beijing},
                country={China}}
\affiliation[3]{organization={Center of Big Data, China Digital Culture Group Co., Ltd},
                postcode={100176}, 
                city={Beijing},
                country={China}}
\affiliation[4]{organization={Department of Stomatology, Tongxiang Hospital of Traditional Chinese Medicine},
                postcode={100024}, 
                city={Tongxiang},
                country={China}}
\affiliation[5]{
    organization={School of Mathematical and Computational Sciences, Massey University}, 
    postcode={100024}, 
    city={Auckland}, 
    country={New Zealand}
}
\affiliation[6]{
    organization={Shining3D Tech Co., Ltd.}, 
    postcode={311258}, 
    city={Hangzhou}, 
    country={China}
}
\affiliation[7]{
    organization={Oral and Maxillofacial Surgery at the Faculty of Dentistry,University of Hong Kong}, 
    postcode={999077}, 
    city={Hong Kong}, 
    country={China}
}
\affiliation[8]{
    organization={School of Art Design, Zhejiang Gongshang University}, 
    postcode={310018}, 
    city={Hangzhou}, 
    country={China}
}
\affiliation[9]{
    organization={Zhejiang Key Laboratory of Big Data and Future E-Commerce Technology}, 
    postcode={310018}, 
    city={Hangzhou}, 
    country={China}
}

%% use optional labels to link authors explicitly to addresses:
%% \author[label1,label2]{}
%% \affiliation[label1]{organization={},
%%             addressline={},
%%             city={},
%%             postcode={},
%%             state={},
%%             country={}}
%%
%% \affiliation[label2]{organization={},
%%             addressline={},
%%             city={},
%%             postcode={},
%%             state={},
%%             country={}}
\cortext[cor1]{Corresponding author}
\fntext[1]{Xinxin Zhao and Jian Jiang contributed equally to this work.}

%% Abstract
\begin{abstract}
Tooth image segmentation is a cornerstone of dental digitization. 
However, traditional image encoders relying on fixed-resolution feature maps often lead to discontinuous segmentation and poor discrimination between target regions and background, due to insufficient modeling of environmental and global context. Moreover, transformer-based self-attention introduces substantial computational overhead because of its quadratic complexity (O(n²)), making it inefficient for high-resolution dental images. To address these challenges, we introduce a three-stage encoder with hierarchical feature representation to capture scale-adaptive information in dental images. By jointly leveraging low-level details and high-level semantics through cross-scale feature fusion, the model effectively preserves fine structural information while maintaining strong contextual awareness. 
Furthermore, a bidirectional sequence modeling strategy is incorporated to enhance global spatial context understanding without incurring high computational cost. 
 We validate our method on two dental datasets, with experimental results demonstrating its superiority over existing approaches. On the OralVision dataset, our model achieves a 1.1\% improvement in mean intersection over union (mIoU).
\end{abstract}

%%Graphical abstract
% \begin{graphicalabstract}
% \includegraphics[width=\textwidth]{method.pdf}
% \end{graphicalabstract}

%%Research highlights
% \begin{highlights}
% \item An effective and efficient dental image segmentation framework is proposed, which achieves high-quality and fine-grained segmentation by enhancing multi-scale representation and contextual modeling, demonstrating superior performance on complex dental images.
% \item A hierarchical feature representation strategy is employed to enable effective multi-scale feature extraction and fusion, which strengthens environmental perception and improves segmentation accuracy, particularly in complex dental imaging scenarios.
% \item A Mamba-based image encoder is developed, integrating a bidirectional sequence block for position-aware sequence modeling, and utilizing Mamba’s selective mechanism to reduce computational complexity while enabling multi-scale feature extraction.
% \end{highlights}

%% Keywords

\end{frontmatter}

%% Add \usepackage{lineno} before \begin{document} and uncomment 
%% following line to enable line numbers
%% \linenumbers

%% main text
%%

%% Use \section commands to start a section
\section{Introduction}
\label{sec:introduction}
% Tooth segmentation, a specialized task in pattern recognition, aims to identify and delineate individual teeth and related anatomical structures from dental images. It holds significant importance in the realm of dental digitization, with possible uses such as diagnosing dental diseases~\citep{chauhan2023overview}, tracking the progress of treatments~\citep{tian2023revised,tian2024rgb}, and conducting image analysis~\citep{yu2024self}. 

Tooth segmentation, a specialized task in pattern recognition, aims to identify and delineate individual teeth and related anatomical structures from dental images. It holds significant importance in the realm of dental digitization, with possible uses such as diagnosing dental diseases~\citep{chauhan2023overview}, tracking the progress of treatments~\citep{tian2023revised,tian2024rgb}, and conducting image analysis~\citep{yu2024self}. 
Recently, foundation segmentation models such as the Segment Anything Model (SAM)~\citep{Kirillov_2023} have shown promising generalization capabilities, motivating their exploration in dental image segmentation.

% challenge

 % In recent years, the Segment Anything Model (SAM) \citep{Kirillov_2023} has attracted considerable attention in the computer vision community due to its strong generalization ability in image segmentation tasks. Several follow-up works \citep{ren2024grounded,xiong2024efficientsam} have focused on improving the efficiency and applicability of SAM in different scenarios. However, directly applying SAM-based models to dental images remains challenging, especially in complex and noisy environments. Although existing improvements \citep{Ma_2024,wu2025medical} have enhanced multi-scale feature capture and fine-grained modeling, they still struggle to effectively represent the relationship between subtle local features and global context, leading to degraded segmentation quality in challenging dental scenarios.

Several recent segmentation approaches \citep{ren2024grounded, xiong2024efficientsam} have focused on improving generalization and efficiency in practical applications. 
However, these methods still face limitations in capturing multi-scale and global contextual information, which hampers their ability to comprehensively model fine-grained details and large-scale contextual relationships. 
This limitation becomes particularly evident when handling complex dental images with noise, where the quality of the generated segmentation masks is often compromised.

Earlier studies \citep{yao2023dual} have primarily focused on the Transformer framework and its self-attention mechanism. However, the quadratic computational complexity of Transformers significantly limits their ability to efficiently process high-resolution dental images. Nevertheless, existing high-quality segmentation methods \citep{liang2023open,Ma_2024} often suffer from slow inference speed and high computational overhead. Therefore, our work aims to develop a segmentation algorithm that achieves high segmentation quality while significantly improving efficiency.

% motivation (how do you get the idea)
% Inspired by the Mamba \citep{zhu2024vision,ma2024u} module and HQ-SAM's \citep{Ke_2023} outstanding performance in their respective fields, this study aims to develop an efficient and high-quality framework for dental image segmentation. Specifically, Mamba's linear complexity 
% in sequence modeling offers a promising solution for reducing computational cost while effectively capturing long-range dependencies. However, as shown in Fig.~\ref{figchallenges}, HQ-SAM struggles with blurry boundaries in specific dental scenarios, making it difficult to accurately segment target regions. Additionally, its limitations in inference speed and computational efficiency highlight the need for further optimization.

\begin{figure}[htbp]
	\centering
	\includegraphics[width=0.85\linewidth]{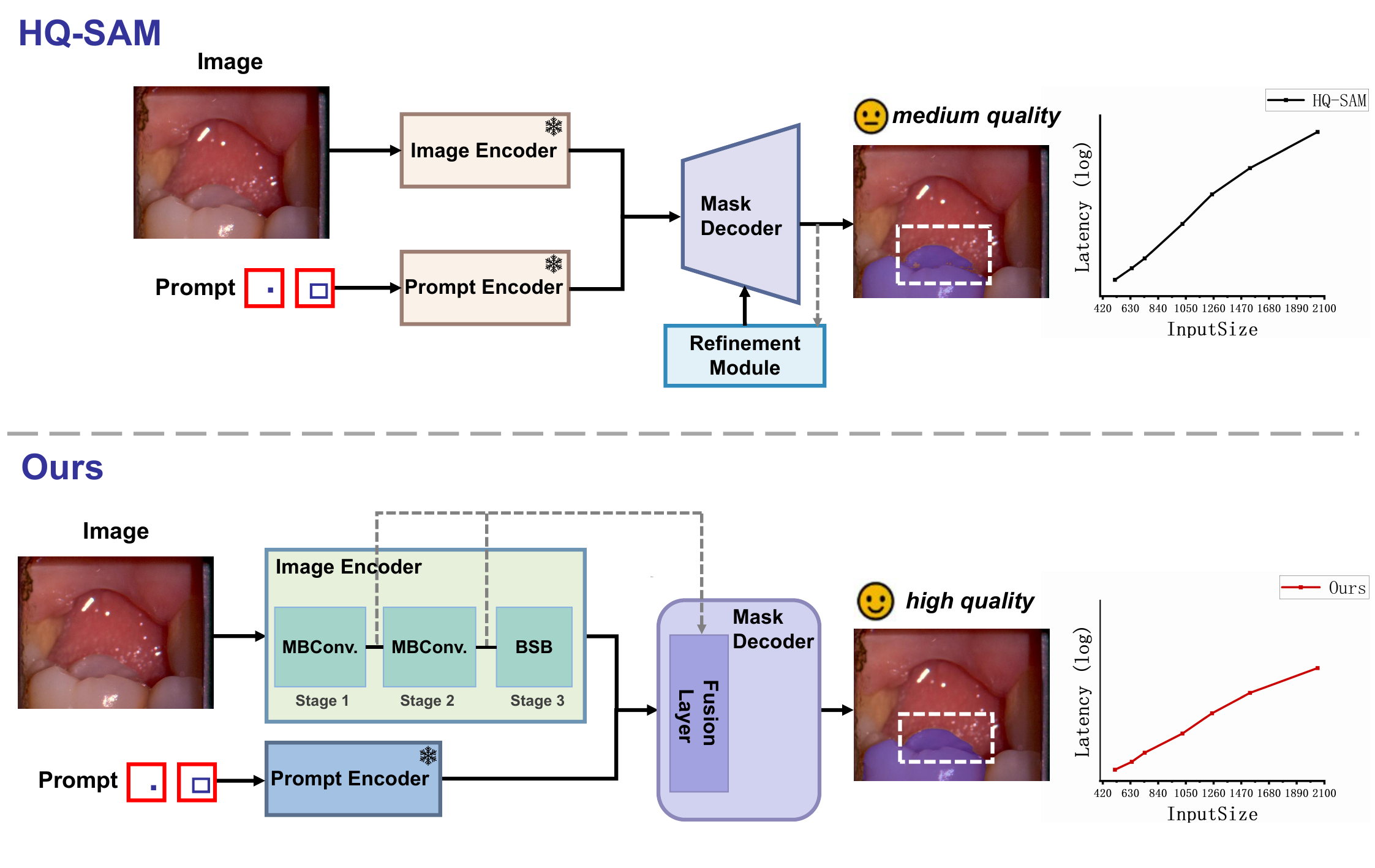} \\
	\caption{\textbf{For the dental segmentation dataset, we conducted a comprehensive comparison between our method and the advanced segmentation approach HQ-SAM in terms of architectural design.} Across image processing tasks with varying resolutions, our method leverages optimized feature extraction strategies and efficient architectural design to maintain low latency while delivering high-quality segmentation results. The blue area represents the mask generated after image segmentation. The white dashed box visually highlights the differences between the segmentation masks generated by our method and those produced by HQ-SAM.}\label{figchallenges}
\end{figure}

Inspired by the linear-complexity sequence modeling capability of Mamba~\citep{zhu2024vision} and the strong mask quality and boundary-aware design of HQ-SAM~\citep{Ke_2023}, this study aims to develop an efficient and high-quality framework tailored for dental image segmentation.
Mamba’s linear computational complexity in sequence modeling makes it a promising candidate for capturing long-range dependencies with reduced computational cost. 
However, directly adopting existing Mamba-based designs in the visual domain does not adequately address the requirements of dental image segmentation, where precise boundary delineation and multi-scale structural details are critical. 
HQ-SAM further demonstrates the effectiveness of enhancing mask quality in SAM-based frameworks by emphasizing fine-grained structures. Nevertheless, as illustrated in Fig.~\ref{figchallenges}, although HQ-SAM produces high-quality masks, it still exhibits blurry boundaries in certain dental scenarios and suffers from limited inference efficiency at high resolutions.
%Fig. 1 illustrate the comparison between different approaches. 

% contribution
% In this study, we build upon the foundational framework of SAM by reusing the prompt encoder and incorporating a three-stage architecture into the image encoder for multi-scale feature extraction. Additionally, we feed the low-level features from the initial two stages into the mask decoder, employing feature fusion to enhance contextual awareness, thereby generating high-quality and fine-grained segmentation masks (as shown in Fig.~\ref{figchallenges}). In comparison with HQ-SAM, our framework incorporates a bidirectional sequence block (BSB) within the image encoder, which facilitates bidirectional feature extraction. This addition improves the model's capability to perceive and analyze environmental details while substantially enhancing the segmentation mask quality. Despite these improvements, the computational demands remain minimal, striking a harmonious balance between the precision of segmentation and efficiency in computation. Unlike previous bidirectional Mamba designs \citep{bimamba,lkm}, our BSB is task-aware and optimized for high-resolution dental image segmentation through lightweight integration. A more in-depth explanation of our approach can be found in Fig.~\ref{method}.

In this study, we extend the SAM framework toward the dental domain by designing a task-oriented image encoder that emphasizes both segmentation quality and computational efficiency. 
The proposed framework introduces a hierarchical representation strategy to well capture the multi-scale characteristics of dental images, enabling more accurate delineation of fine anatomical structures. 
To further enhance global contextual understanding, a bidirectional sequence block (BSB) is incorporated into the encoder, allowing information to be aggregated from complementary spatial directions. 
Compared with HQ-SAM (as shown in Fig.~\ref{figchallenges}), this design improves the perception of complex oral environments while maintaining efficient inference. 
Unlike previous bidirectional Mamba designs \citep{bimamba,lkm}, our BSB is task-aware and optimized for high-resolution dental image segmentation through lightweight integration. A more in-depth explanation of our approach can be found in Fig.~\ref{method}.

% highlight
The main contributions of this paper are outlined below:
% \begin{itemize}
% 	\item An effective framework for segmenting dental images is presented, employing a three-stage architecture that facilitates the efficient extraction of multi-scale features, thereby significantly enhancing the model’s ability to discern intricate details and contextual nuances within complex dental images.
% 	\item Multi-scale feature extraction and feature fusion are leveraged to strengthen environmental perception and improve segmentation accuracy, particularly in complex dental images.
% 	\item A Mamba-based image encoder is developed, integrating a bidirectional sequence block for position-aware sequence modeling, and utilizing Mamba’s selective mechanism to reduce computational complexity while enabling multi-scale feature extraction.
% \end{itemize}
\begin{itemize}
    \item An effective and efficient dental image segmentation framework is proposed, which achieves high-quality and fine-grained segmentation by enhancing multi-scale representation and contextual modeling, demonstrating superior performance on complex dental images.
    \item A hierarchical feature representation strategy is employed to enable effective multi-scale feature extraction and fusion, which strengthens environmental perception and improves segmentation accuracy, particularly in complex dental imaging scenarios.
    \item A Mamba-based image encoder is developed, integrating a bidirectional sequence block for position-aware sequence modeling, and utilizing Mamba’s selective mechanism to reduce computational complexity while enabling multi-scale feature extraction.
\end{itemize}

% experimental results
The experimental results on the dental segmentation dataset (DSD) \citep{TIAN2020107158} and OralVision \citep{OralVision} dataset demonstrate that our method effectively generates high-quality segmentation masks while efficiently handling the challenges posed by high-resolution input images. Even at higher image resolutions, our approach maintains lower latency compared to other methods, showcasing the balance between performance and efficiency.

\section{Related Work}\label{relatedWork}
We provide a concise overview of recent studies on high-quality segmentation, zero-shot image segmentation, linear attention models, and multi-scale feature extraction.

\subsection{High-quality Segmentation}
High-quality image segmentation aims to precisely delineate specific regions within an image~\citep{tian2023survey}. 
The use of high-quality segmentation techniques has become prevalent across a range of applications, including semantic segmentation~\citep{jiang2025cgvit}, instance segmentation~\citep{ouyang2024mixingmask}, and panoptic segmentation~\citep{chu2022learning}, as shown in numerous studies~\citep{tian20223d}. Traditional approaches~\citep{qian2023automatic}, especially those leveraging convolutional neural networks (CNNs), have been widely adopted. Nevertheless, these strategies frequently encounter challenges in effectively capturing long-range dependencies and contextual nuances, prompting researchers to investigate more advanced architectures.
The introduction of Transformers provided a new approach to addressing these challenges.
Mask Transformer~\citep{cheng2022masked} has successfully employed Transformer-based architectures to address the limitations of CNNs in high-quality segmentation tasks. For instance, Segmenter~\citep{strudel2021segmenter} proposed a Transformer architecture specifically designed for semantic segmentation. By employing an effective self-attention mechanism and positional encoding, it significantly enhances the accuracy and efficiency of semantic segmentation.
To further unify the image segmentation framework, Mask2Former~\citep{cheng2022masked} introduced a masked-attention mechanism. This approach processes image regions selectively and combines a shared encoder with task-specific decoders to handle various segmentation tasks, substantially improving segmentation precision.
Moreover, the growing demand for real-time, high-quality segmentation on mobile devices has brought increased attention to mobile segmentation. 
MobileFormer~\citep{chen2022mobile} combines the lightweight MobileNet with Transformer architecture, proposing an efficient model tailored for high-quality image segmentation tasks on mobile devices. However, these methods overlook the issue that the self-attention mechanism of Transformers has high computational complexity in long-sequence tasks, particularly in high-resolution images.

\subsection{Zero-shot Image Segmentation}
Zero-shot image segmentation seeks to accurately segment specific categories within an image without having seen samples of these categories during training. Liang et al.~\citep{liang2023open} have leveraged large-scale image-text pairs for pre-training, endowing models with strong image understanding and text alignment capabilities, thereby enhancing their ability to segment unseen categories. The SAM~\citep{Kirillov_2023} further advances this capability by utilizing large-scale pre-training, multimodal learning, contrastive learning, and flexible input prompts, achieving zero-shot segmentation on unseen images.
Some studies~\citep{Ke_2023,Ma_2024} have built upon SAM as a foundational model to enhance zero-shot segmentation in specific domains. For example, Ma et al. proposed MedSAM~\citep{Ma_2024}, which maintains the core SAM architecture but is trained on an extensive medical image dataset. They implemented necessary modifications and enhancements to enable zero-shot performance in medical image segmentation tasks. To enhance environmental perception and adapt to new, unseen tasks and complex environments, zero-shot learning enables models to generalize from previously acquired knowledge. This capability helps models understand and process diverse inputs without requiring direct exposure to the specific data, thereby improving performance in challenging scenarios.

\subsection{Linear Attention Model}
The emergence of the linear attention model mainly tackles the shortcomings of conventional attention models, such as those found in Transformers, in terms of computational complexity and memory usage. Mamba~\citep{gu2023mamba} and RWKV~\citep{peng2023rwkv,duan2024vision} present advantages over Transformers in computational efficiency and memory consumption. Mamba achieves reduced computational complexity by optimizing the attention mechanism, while RWKV significantly reduces memory usage by incorporating RNN characteristics and improving memory management. Duan et al.~\citep{duan2024vision} applied RWKV to the visual domain, where VRWKV outperformed ViT~\citep{yao2023dual} in image processing performance and demonstrated significantly faster speeds and lower memory usage when handling high-resolution inputs. Another significant contribution is vision mamba (Vim)~\citep{zhu2024vision}, which integrates bidirectional SSM~\citep{lu2024structured} for comprehensive global visual context modeling, along with position embeddings that facilitate a position-aware interpretation of visuals. Nonetheless, these methods often overlook the importance of multi-scale feature representation in visual tasks.

\begin{figure}
	\centering
	{\includegraphics[width=.85 \textwidth]{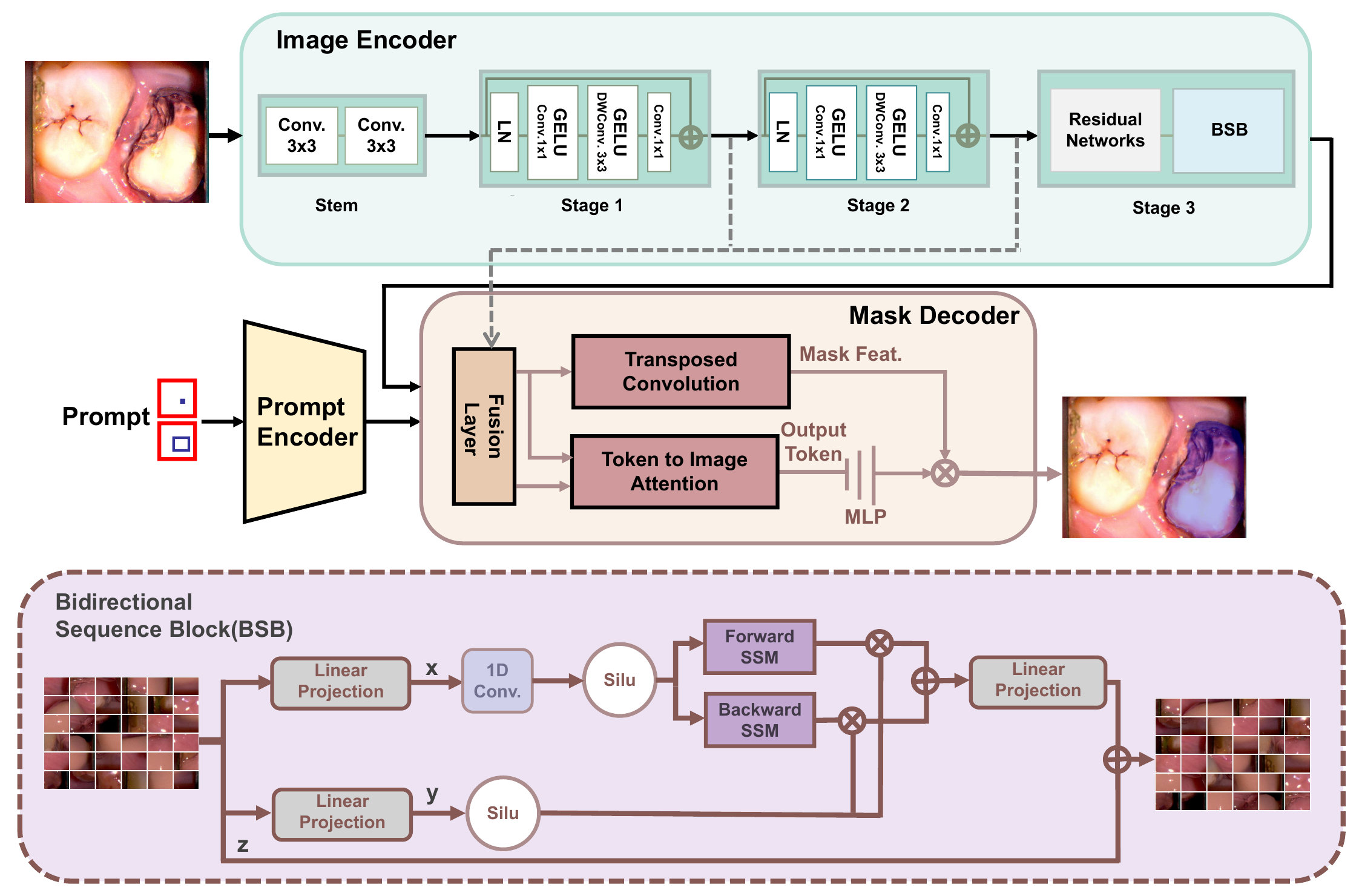}}\\ 
	\caption{\textbf{Overall structure of the proposed approach.}
		The proposed approach adopts a classic encoder-decoder architecture. First, the input dental images undergo a three-stage downsampling process to capture image features at multiple scales. These extracted features are then combined with prompt vectors created by the prompt encoder and processed together by the mask decoder to generate segmentation masks that correspond to the original images. In BSB, the sigmoid linear unit (Silu) activation function combines the smoothness of the sigmoid with a linear component to enhance feature representation. The blue region indicates the mask generated following the image segmentation process.
	}\label{method}
\end{figure}
\subsection{Multi-scale Feature Extraction}
Multi-scale feature extraction plays a crucial role in dense prediction tasks by simultaneously capturing fine-grained details and global contextual information. Traditional convolutional networks often face a trade-off between accuracy and efficiency when handling multi-scale features. To overcome this limitation, recent studies have extensively explored hybrid architectures and efficient feature fusion mechanisms. ViT-CoMer~\citep{xia2024vit} embeds convolutional modules into a Transformer-based architecture, achieving more effective multi-scale feature interaction while preserving the advantage of a global receptive field. In the domains of remote sensing and geological hazard analysis, multi-scale attention mechanisms have also demonstrated strong segmentation performance under complex textures. Cai et al.~\citep{cai2024multi} and Yang et al.~\citep{yang2025state} proposed attention-based multi-scale modeling approaches that significantly enhance the model's ability to perceive subtle regional differences in geological images. In addition, Jiang et al.~\citep{jiang2024mffsodnet} and Li et al.~\citep{li2024mffsp} designed multi-scale fusion strategies for small object detection and scene parsing, respectively, which improved model performance in high-resolution and noisy environments. Despite the remarkable progress achieved in these areas, most existing methods remain focused on natural or remote sensing images. There is still a lack of targeted exploration in specialized domains such as dental imagery, where both structural detail and inference efficiency are critical.

\section{Our Approach} \label{ourApproach}
We propose a dental image segmentation method based on the SAM model architecture. This method generates corresponding segmentation masks by inputting dental images along with point or box prompts. As shown in Fig.~\ref{method}, our method employs a classical encoder-decoder framework. Initially, dental images are processed through a three-stage downsampling pipeline in the encoder. In the third stage, a bidirectional sequence block performs forward and backward scanning over patch blocks to capture global contextual features. These extracted features are then integrated with prompt embeddings produced by the prompt encoder. The integrated information is then processed by the mask decoder, producing segmentation masks for the input images.

\subsection{Preliminaries: State Space Model}
State space models (SSM) were originally designed to handle time series or data with causal relationships, which makes their direct application to 2D image data non-trivial. In this case, images are typically unrolled into sequences, and scanning ensures that all pixel information is captured, enabling the conversion of spatial features into visual features. The SSM is a framework for describing time series data, commonly used for handling dynamic systems and time-varying processes. SSM illustrates how system states change over time and the connection between the observed data and these states. The model is represented by a hidden latent state $\mathbf{h}(t) \in \mathbb{R}^{N}$, mapping a one-dimensional sequence $\mathbf{x}(t) \in \mathbb{R}$ $\mapsto$ $\mathbf{y}(t) \in \mathbb{R}$. This process can be expressed by the following linear ordinary differential equations:
\begin{eqnarray}
	\mathbf{h}'(t) &=& \mathbf{A}\mathbf{h}(t) + \mathbf{B}\mathbf{x}(t),\\
	\mathbf{y}(t) &=& \mathbf{C}\mathbf{h}(t) + \mathbf{D}\mathbf{x}(t).
\end{eqnarray}
where the parameter matrices $\mathbf{A} \in \mathbb{R}^{N \times N}$, $\mathbf{B} \in \mathbb{R}^{N \times 1}$, $\mathbf{C} \in \mathbb{R}^{1 \times N}$, and $\mathbf{D} \in \mathbb{R}$ define the relationships between the system's states, inputs, and outputs.

Within the Mamba framework, the continuous-time state space model is discretized by introducing a sampling interval $\Delta t$. Specifically, the continuous parameters $\mathbf{A}$ and $\mathbf{B}$ are transformed into their discrete counterparts $\mathbf{A}_d$ and $\mathbf{B}_d$ through the zero-order hold (ZOH) method:
\renewcommand{\arraystretch}{1.2} % 调整行距
\setlength\arraycolsep{2pt}       % 调整列间距
\begin{eqnarray}
\mathbf{A}_d &=& \exp(\mathbf{A}\Delta t), \\
\mathbf{B}_d &=& \left(\int_0^{\Delta t}\exp(\mathbf{A}\tau)\,d\tau\right)\mathbf{B}.
\end{eqnarray}

The discretized representation of this system can be stated as:
\begin{eqnarray}
	\mathbf{h}_t &=& \mathbf{A}_d \mathbf{h}_{t-1} + \mathbf{B}_d \mathbf{x}_t,\\
	\mathbf{y}_t &=& \mathbf{C}\mathbf{h}_t + \mathbf{D}\mathbf{x}_t.
\end{eqnarray}

Mamba, evolving from the standard SSM, modifies the original linear constraints by rendering $\boldsymbol{\beta}$ and $\boldsymbol{\theta}$ contingent on the input. This adaptation allows Mamba to uphold linear computational complexity throughout the forward propagation phase. For simplicity and computational efficiency, we set $\mathbf{D} = 0$ in our implementation.

\subsection{The bidirectional sequence block}
Transformer-based image segmentation methods require processing the entire image representation in each layer to capture global attention, which significantly increases the computational burden. Moreover, when dealing with high-resolution images, the input must be divided into more patches, leading to a substantial increase in sequence length and resulting in a computational complexity of $O(n^2)$. To reduce computational costs, we draw inspiration from the Mamba framework. While vanilla Mamba blocks reduce computational complexity, their unidirectional scanning mechanism may limit the propagation of early visual information. Additionally, Mamba's position-awareness capability in vision tasks is limited. 

\begin{figure}[htbp]
	\centering
	\includegraphics[width=0.7\linewidth]{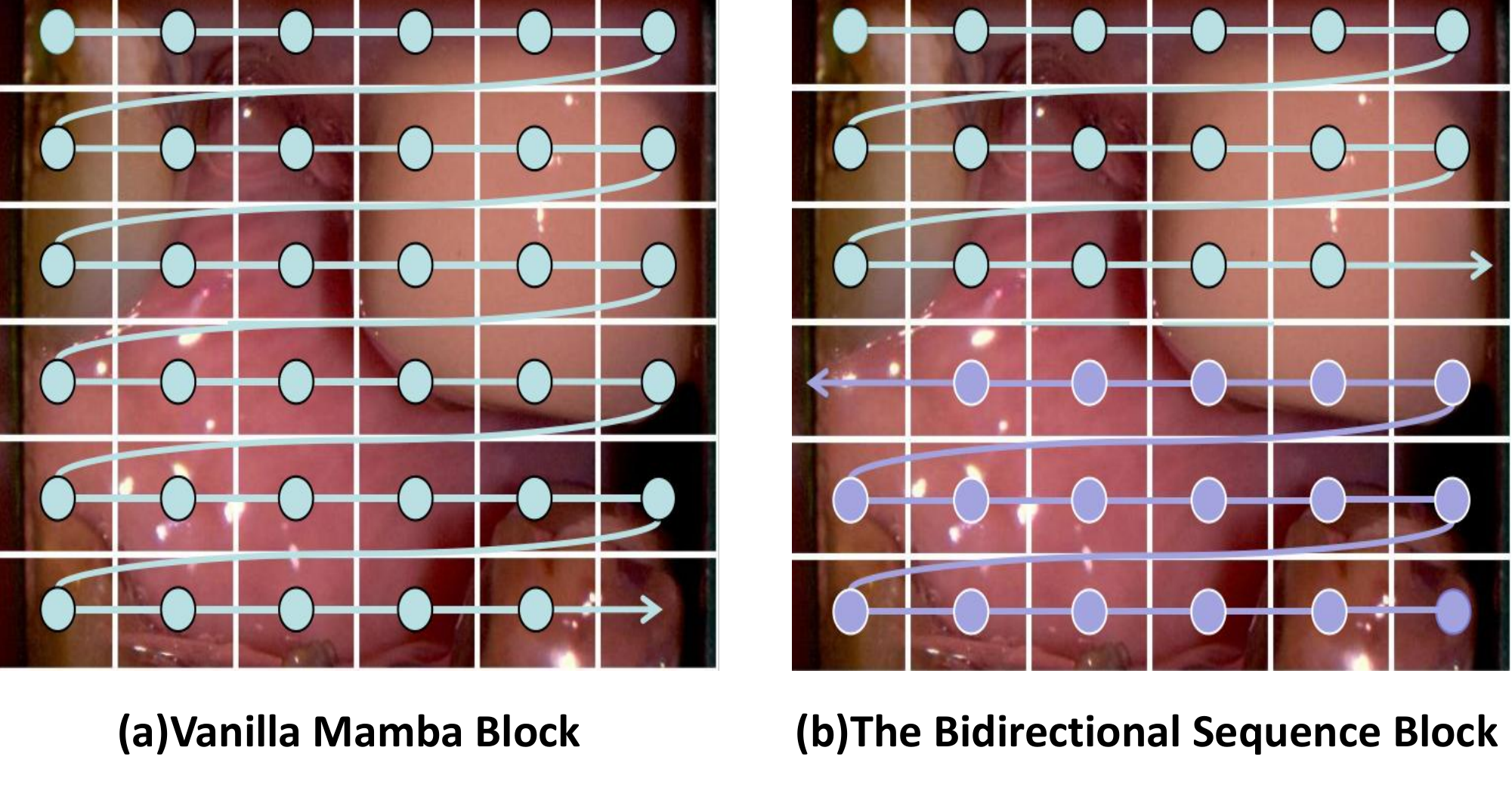}\\
	\caption{\textbf{Illustration of the comparison between the vanilla mamba block and the bidirectional sequence block.} 
		(a) Vanilla mamba blocks are scanned in a sequential order from start to finish.
		(b) The bidirectional sequence module scans in a forward and backward order.
		The green and purple dots represent patch blocks at different positions, while the arrows in different colors indicate various scanning orders.}\label{figScan}
\end{figure}

To address these issues, we introduce a bidirectional sequence block into the Mamba-based architecture, enabling the consideration of both forward and backward visual contexts. As shown in Fig.~\ref{figScan}, we illustrate the differences in scanning sequences between the two methods. Our approach performs both forward and backward scanning simultaneously and integrates the outputs, thereby capturing more comprehensive contextual information from the image. Algorithm \ref{alg:algorithm1} summarizes the implementation process of the bidirectional sequence block. 

\begin{algorithm}
\small
	\caption{BSB Process}  \label{alg:algorithm1}
	\textbf{Input:} token sequence $\mathbf{P}_{l-1}$;\\
	\textbf{Output:} token sequence $\mathbf{P}_l$;\\
    \textbf{Definition:} $p_i$ denotes the learnable continuous-time state parameter; \\
	$\mathbf{P}^{\prime}_{l-1} = \mathrm{Norm}(\mathbf{P}_{l-1})$;\\
	$\mathbf{x} = \mathrm{Linear}_x(\mathbf{P}^{\prime}_{l-1})$;\\
	$\mathbf{y} = \mathrm{Linear}_y(\mathbf{P}^{\prime}_{l-1})$;\\
	$\mathbf{z} = \mathbf{P}_{l-1}$;\\

	\For{$i \in \{\text{forward}, \text{backward}\}$}{
		$\mathbf{x}^{\prime}_{i} = \mathrm{SiLU}(\mathrm{Conv1d}_{i}(\mathbf{x}))$;\\
		$\mathbf{B}_{i} = \mathrm{Linear}^{B}_{i}(\mathbf{x}^{\prime}_{i})$;\\
		$\mathbf{C}_{i} = \mathrm{Linear}^{C}_{i}(\mathbf{x}^{\prime}_{i})$;\\
		$\Delta_{i} = \log(1+\exp(\mathrm{Linear}^{\Delta}_{i}(\mathbf{x}^{\prime}_{i})+p_{i}))$;\\
		$\overline{\mathbf{A}}_{i} = \Delta_{i}\otimes p_{i}$;\\
		$\overline{\mathbf{B}}_{i} = \Delta_{i}\otimes \mathbf{B}_{i}$;\\
		$\mathbf{k}_{i} = \mathrm{SSM}(\overline{\mathbf{A}}_{i},\overline{\mathbf{B}}_{i}, \mathbf{C}_{i})(\mathbf{x}^{\prime}_{i})$;\\
	}
	$g_{\text{forward}} = \mathrm{SiLU}(\mathrm{Linear}^{g}_{\text{fwd}}(\mathbf{y}))$; \\
    $g_{\text{backward}} = \mathrm{SiLU}(\mathrm{Linear}^{g}_{\text{bwd}}(\mathbf{y}))$; \\
    $\mathbf{k}^{\prime}_{\text{forward}} = \mathbf{k}_{\text{forward}} \odot g_{\text{forward}}$;\\
    $\mathbf{k}^{\prime}_{\text{backward}} = \mathbf{k}_{\text{backward}} \odot g_{\text{backward}}$;\\
	$\mathbf{P}_{l} = \mathrm{Linear}(\mathbf{k}^{\prime}_{\text{forward}}+\mathbf{k}^{\prime}_{\text{backward}}) + \mathbf{z}$;\\
\end{algorithm}

In particular, the input sequence $\mathbf{P}_{l-1}$ is first normalized to obtain $\mathbf{P}^{\prime}_{l-1}$. From this normalized representation, two branches are derived: one projected into $\mathbf{x}$, which is further processed by a one-dimensional convolution and nonlinearity, and the other projected into $\mathbf{y}$, which serves as a gating signal. Meanwhile, we explicitly define $\mathbf{z}$ which represents the residual connection carried from the block input. The processed $\mathbf{x}$ is passed through forward and backward SSMs to obtain $\mathbf{k}^{\prime}_{\text{forward}}$ and $\mathbf{k}^{\prime}_{\text{backward}}$. These are modulated by the gate from $\mathbf{y}$, fused, and finally added to the residual $\mathbf{z}$ to produce the updated sequence:
\begin{eqnarray}
	\mathbf{P}_{l} = \mathrm{BSB}(\mathbf{P}_{l-1}) + \mathbf{z}, \label{eq7}
\end{eqnarray}
where BSB is the bidirectional sequence block, with $l$ indicating the layer number. 

The BSB replaces the quadratic self-attention with a linear-complexity state-space model and introduces two direction-specific gating mechanisms to adaptively fuse the forward and backward branches. Unlike a shared modulation, the separate gates $g_{\text{forward}}$ and $g_{\text{backward}}$ reweight their respective branches independently, which helps emphasize structure-relevant features while reducing redundant responses. This design contributes to more stable feature modeling and improved precision in fine-grained dental segmentation.

To mitigate the limitations of applying 1D SSMs to 2D data, the BSB operates only on downsampled feature maps with compact spatial size. 
Combined with lightweight convolution and gated fusion, this design preserves essential spatial dependencies without requiring explicit positional encodings. 
Instead of naïve flattening, the feature map is partitioned into non-overlapping $m\times n$ sub-kernels and serialized within each sub-kernel in raster order, which maintains local continuity. 
Stacked $3\times3$ convolutions, together with pooling and gated fusion, further enlarge the receptive field beyond local neighborhoods, while the bidirectional SSM stage propagates longer-range dependencies.

\textbf{2D to 1D conversion.} 
To transform a 2D feature map into a 1D sequence, 
we first partition the feature map into non-overlapping sub-kernels of size $m\times n$. 
Within each sub-kernel, pixels are serialized in raster order, preserving local continuity 
and avoiding the loss of neighborhood information caused by naïve flattening. 
Each serialized sequence is then processed by the state-space model in both forward and backward directions. 
The bidirectional scan enables tokens to aggregate context from both preceding and succeeding positions. 
Finally, pooled sub-kernel representations are aggregated to capture global dependencies and 
unpooled back to the original resolution, ensuring that both local and global spatial contexts are effectively modeled.

\subsection{The Efficient Segmentation Model}
In previous research, the lack of effective representation of fine-grained features in complex oral images often led to suboptimal dental image segmentation performance. Moreover, conventional techniques focused mainly on local receptive fields, limiting their ability to grasp long-range dependencies and overarching contextual details. Although dividing images into fixed-size patches and using self-attention mechanisms can capture some information, these methods struggle to effectively extract multi-scale features due to the absence of explicit multi-scale modeling.

Inspired by the feature pyramid structure, we integrated it into the encoder and utilized the richer spatial, texture, and other detailed features generated during encoding to guide finer predictions in the decoding phase. This method tackles the problem of information loss that often arises from depending only on high-level features. Moreover, by incorporating multi-scale features, the model becomes more attuned to segmentation targets within the dental imaging context, greatly boosting its robustness.

\textbf{Fusion layer.} In the decoder, feature fusion is implemented in a top-down manner similar to feature pyramid networks. Specifically, the $16\times$ stage feature $\mathbf{F}$ is first upsampled by a factor of 2 (bilinear interpolation) to match the spatial size of the $8\times$ feature $\mathbf{F}_{mf}$. The two feature maps are then concatenated along the channel dimension, followed by a $1\times1$ convolution to unify the channel width (128 in our implementation). The fused feature is further upsampled by another factor of 2 and concatenated with the $4\times$ feature $\mathbf{F}_{lf}$, again reduced by a $1\times1$ convolution to 128 channels. Finally, the fused feature passes through a lightweight refinement block. This hierarchical fusion progressively restores spatial resolution while integrating semantic context from deeper stages.

To enhance the precision of dental image segmentation, we developed a three-tiered feature pyramid architecture within the image encoder, aiming to enrich visual context and accurately capture intricate low-level details. Initially, the input image $\mathbf{I}$ undergoes preliminary feature extraction and downsampling. In the first two stages, we perform $4\times$ and $8\times$ downsampling to extract low-level features of dental images, defined as:
\begin{eqnarray}
	\mathbf{F}_{lf} &=& \mathrm{DownSampling}(\mathbf{I}), \\
	\mathbf{F}_{mf} &=& \mathrm{DownSampling}(\mathbf{F}_{lf}).
\end{eqnarray}

Both stages utilize convolutional blocks to produce higher-resolution feature maps relative to the $16\times$ stage. In crafting the convolution module, we take a cue from the ViTamin framework \citep{chen2024vitamin}, initiating the process with layer normalization (LN) as the foundational element. We start with a $1\times1$ convolution to broaden the channel dimensions, which boosts the model's ability to represent features. Next, a $3\times3$ convolution is implemented for spatial mixing, enhancing the model's overall global perception. Following that, we apply another $1\times1$ convolution for linear projection, which brings the channels back to their original dimensions. Ultimately, the transformed features are integrated with the input features through element-wise addition, allowing for effective feature fusion while preserving the integrity of the information.

In the third stage, we introduce the bidirectional sequence block to obtain global features through $16\times$ downsampling, defined as:
\begin{eqnarray}
	\mathbf{F} &=& \mathrm{BSB}(\mathbf{F}_{mf}),
\end{eqnarray}
where $\mathbf{F}$ is the feature map generated at this stage. 

To enhance the mask characteristics, the decoder integrates low-level features extracted during the initial two stages. The refinement process is carried out using the decoder $\Phi_{\text{dec}}$ and the prompt encoder $\Phi_{\text{prompt}}$, defined as 
\begin{eqnarray}
	\mathbf{M} &=& \Phi_{\text{dec}}(\mathbf{F}, \mathbf{F}_{lf}, \mathbf{F}_{mf}, \Phi_{\text{prompt}}(\mathbf{p})),
\end{eqnarray}
where $\mathbf{p}$ represents the input visual prompt. In our setting, each image is provided with one positive point prompt per annotated object and an optional bounding box prompt during training. Positive points are randomly sampled from the interior of the ground-truth mask, while negative points are randomly sampled from the background regions. To effectively merge these low-level features, we employ a fusion layer for efficient feature fusion, enabling precise dental image segmentation.

The model is designed to produce multi-scale feature maps rather than depending on a singular scale, thus capturing richer details and significantly enhancing segmentation performance. We adopted a combination of softmax cross-entropy and Dice loss. Specifically, the softmax function was applied over all classes, and the Dice loss was computed per class and averaged. This design ensures stable optimization while balancing class imbalance.

\section{Results}  \label{results}
We assess the efficacy of our method and compare it with other approaches on two datasets.

\subsection{Datasets and Evaluation Criteria}
\subsubsection{Datasets}
We evaluated an assessment of the suggested image segmentation method using two distinct dental datasets, namely the Dental Segmentation Dataset \citep{TIAN2020107158} and the OralVision \citep{OralVision} dataset. The following are the details of two dental datasets.

\textbf{Dental Segmentation Dataset}~\citep{TIAN2020107158} comprises a training set of 4900 samples, along with validation and test sets containing 151 samples each. Each sample is accompanied by pixel-level annotations. Each sample is accompanied by pixel-level annotations. In our training pipeline, we did not apply noise or rotation augmentations. We only used standard transformations such as flipping, cropping, and color jittering. 

\textbf{OralVision}~\citep{OralVision} consists of intraoral images captured by dental scanning devices, with a resolution of 640×480. These images were collected from random patients at dental clinics and were annotated pixel by pixel by six researchers from a university using LabelM. The dental images were categorized into tissue types such as gums, lips, teeth, tongue, and cheeks. Each image contains fine-grained segmentation masks and includes noise such as food debris, dental calculus, implant attachments, and saliva. The OralVision dataset comprises a training set, validation set, and test set with 4000, 120, and 120 samples, respectively.

The use of the DSD and OralVision datasets strictly adhered to their respective release terms. For OralVision, written informed consent was obtained from all participants. All images were fully de-identified prior to annotation: direct identifiers and metadata were removed, and each image was assigned a random code. For both datasets, all segmentation categories are present in both training and testing sets, and the training and testing images come from different patients, ensuring unbiased evaluation and preventing potential data leakage.

\subsubsection{Evaluation Criteria}
Based on previous image segmentation methods, we utilized mean Intersection over Union (mIoU) and boundary mean IoU (mBIoU) as evaluation metrics. These metrics effectively measure the model's segmentation performance and boundary precision when handling unseen categories. IoU metrics were computed in a per-class manner and then averaged to obtain mIoU and mBIoU, reflecting performance across all tissue categories.

We compute boundary IoU following common practice: for each class, compute the boundary region by extracting the class contour and dilating it by a tolerance $t$ pixels (we use $t=3$). Let $B_{gt}$ and $B_{pred}$ be the dilated boundary pixels of ground-truth and prediction respectively; the boundary IoU for the class is $\mathrm{IoU}_{\text{boundary}} = |B_{gt}\cap B_{pred}| / |B_{gt}\cup B_{pred}|$.

We also report three efficiency metrics: frames per second (FPS), memory usage (Memory), and floating point operations (FLOPs). FPS and memory were measured during inference with a batch size of 1 and an input resolution of 640×480. FLOPs were computed with fvcore on an input size of 640×480 with a single bounding box prompt, covering both encoder and decoder.

\subsection{Implementation Details}
We conducted experiments on a workstation equipped with an Intel Core i5-13900KF processor (24 cores, base clock 3.5 GHz), 32 GB of RAM, and 4 NVIDIA RTX 4090D GPUs. All models were implemented in PyTorch with cuDNN enabled. Training was performed using 4 GPUs, while inference efficiency metrics (FPS and Memory) were measured on a single RTX 4090D GPU.

The training batch sizes were set at 32 for the DSD dataset and 24 for OralVision, using SAM as the core framework. For optimization, the AdamW algorithm was utilized, featuring a base learning rate of 6e-5 and a weight decay factor of 0.01. Training was run for up to 45 epochs with early stopping based on validation mIoU (patience = 10). The best checkpoints were reached at 31.8 ± 4.6 epochs on DSD and 29.5 ± 5.2 epochs on OralVision. During testing, we applied test-time augmentation (TTA) consisting of horizontal/vertical flips and rotations. Predictions from augmented inputs were transformed back and averaged at the probability level. To ensure fairness, the same TTA procedure was applied consistently across all compared methods.

Normalization was block-specific: LayerNorm (eps = 1e-6) was used in Bidirectional SSM blocks, while BatchNorm2d (eps = 1e-5, momentum 0.1) was used in convolutional stems and upsampling layers. Newly added convolutional layers were initialized with Kaiming normal, linear layers with Xavier uniform, and all biases to zero. Dropout ($p=0.1$) was applied in decoder projections, and stochastic depth (drop-path) with maximum rate 0.1 increased linearly.

We used softmax cross-entropy combined with a multi-class Dice loss, with label smoothing $\varepsilon=0.1$. 
To mitigate class imbalance, class weights were set as $w_c \propto (\mathrm{freq}_c)^{-0.5}$ and normalized. 
At inference, predictions were generated by taking the softmax argmax per pixel without any additional post-processing.

During inference, all SAM-based models (including ours) were evaluated with YOLOv10-generated~\citep{wang2024yolov10} bounding box prompts, using the identical prompts across methods. Non-SAM models (e.g., SegFormer, Swin-Unet, Vim, VRWKV, U-Mamba) operated without prompts. This respects each model’s native setting and ensures a fair comparison under consistent datasets and metrics.

To assess robustness and generalizability, we performed five-fold cross-validation within the training set of both datasets, while keeping the official validation and test sets strictly held out. In each fold, the training portion was further split into four folds for training and one for validation, and the model was trained and evaluated accordingly. Hyperparameters were selected based on cross-validation results, and final performance was reported on the official test sets only. This ensures that the test data remained completely unseen during training and model selection, avoiding any risk of leakage.

\subsection{Ablation Study}
To assess the contribution of each module, we conducted ablation studies on the dental segmentation dataset to analyze how different design choices influence segmentation performance.

\textbf{Backbone.} To validate the efficacy of the encoder in our method, we selected the SAM \citep{Kirillov_2023} model as the backbone for ablation studies. As a powerful image segmentation model, SAM demonstrates exceptional performance in multi-scale feature extraction, providing high-quality image feature representations. In the ablation study, we replaced the feature extraction module in our method with SAM's encoder while keeping the decoder and prompt encoder modules unchanged. This setup allowed us to analyze the impact of our encoder design and the bidirectional sequence block on the overall segmentation performance. By comparing the experimental results, we analyze the effect of the three-stage feature pyramid structure and the bidirectional sequence block on fine-grained feature extraction and segmentation in complex oral environments.

\begin{table}[htbp]
	\centering
    \small
	\begin{tabular}{lc}
		\toprule
		\textbf{Bidirectional Strategy} & \textbf{DSD(mIoU)} \\ 
		\midrule
		None                             & 89.1  $\pm$ 0.4                      \\ 
		\midrule
		Bidirectional SSM                & 90.7    $\pm$ 0.3                    \\ 
		Bidirectional SSM + Conv1d       & \cellcolor{blue!10}\textbf{90.9$\pm$ 0.3}                         \\ 
		\bottomrule
	\end{tabular}
    \caption{\textbf{Ablation study on the bidirectional design. We highlighted our block configuration in blue.}}
    \label{tab:bidirectional_comparison}
\end{table}

\begin{figure*}[htbp]
	\centering
	\subfigure[]{
		\includegraphics[width=0.27 \linewidth]{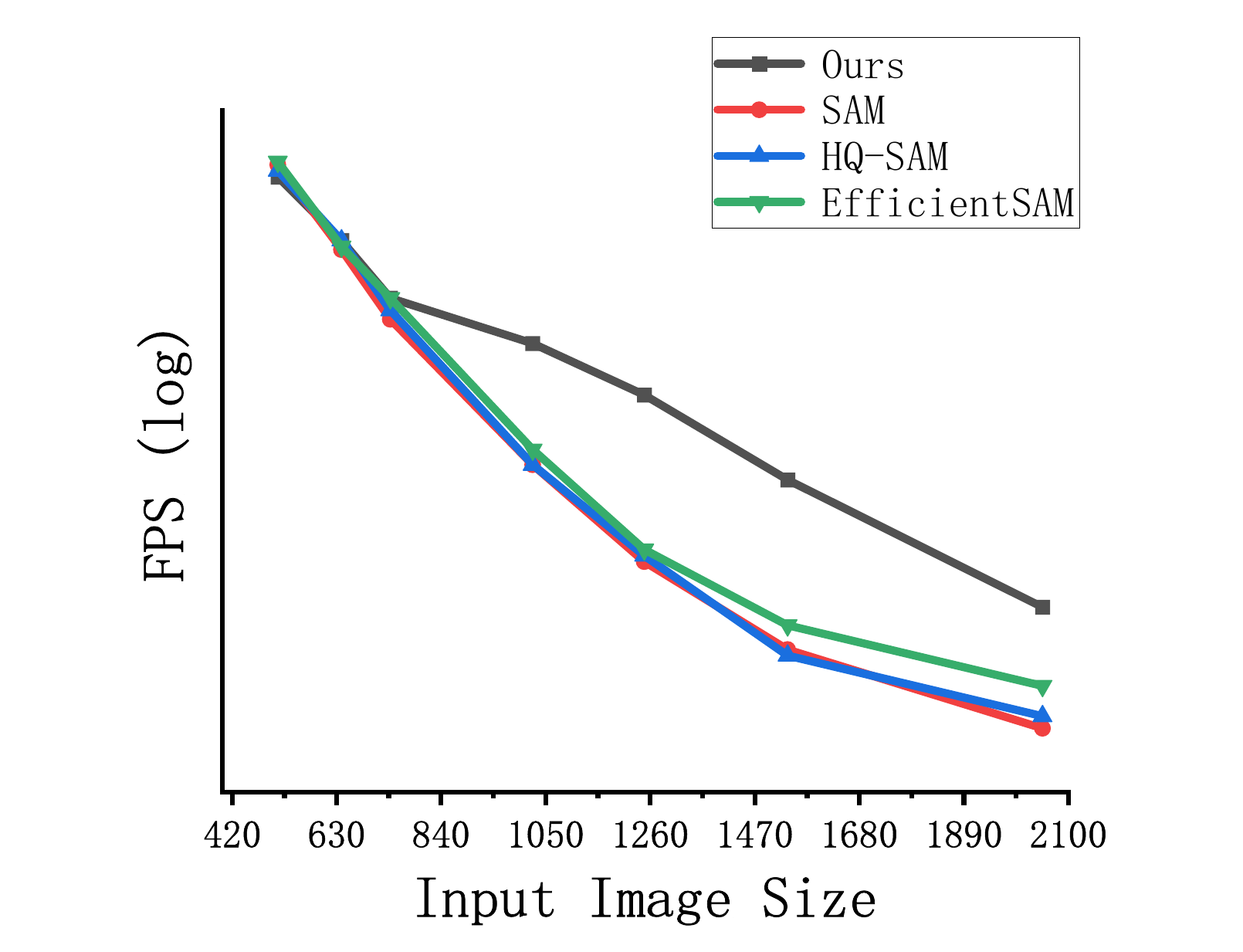}                                                        
	}\hspace{0.02\linewidth}
	\subfigure[]{
		\includegraphics[width=0.14 \linewidth]{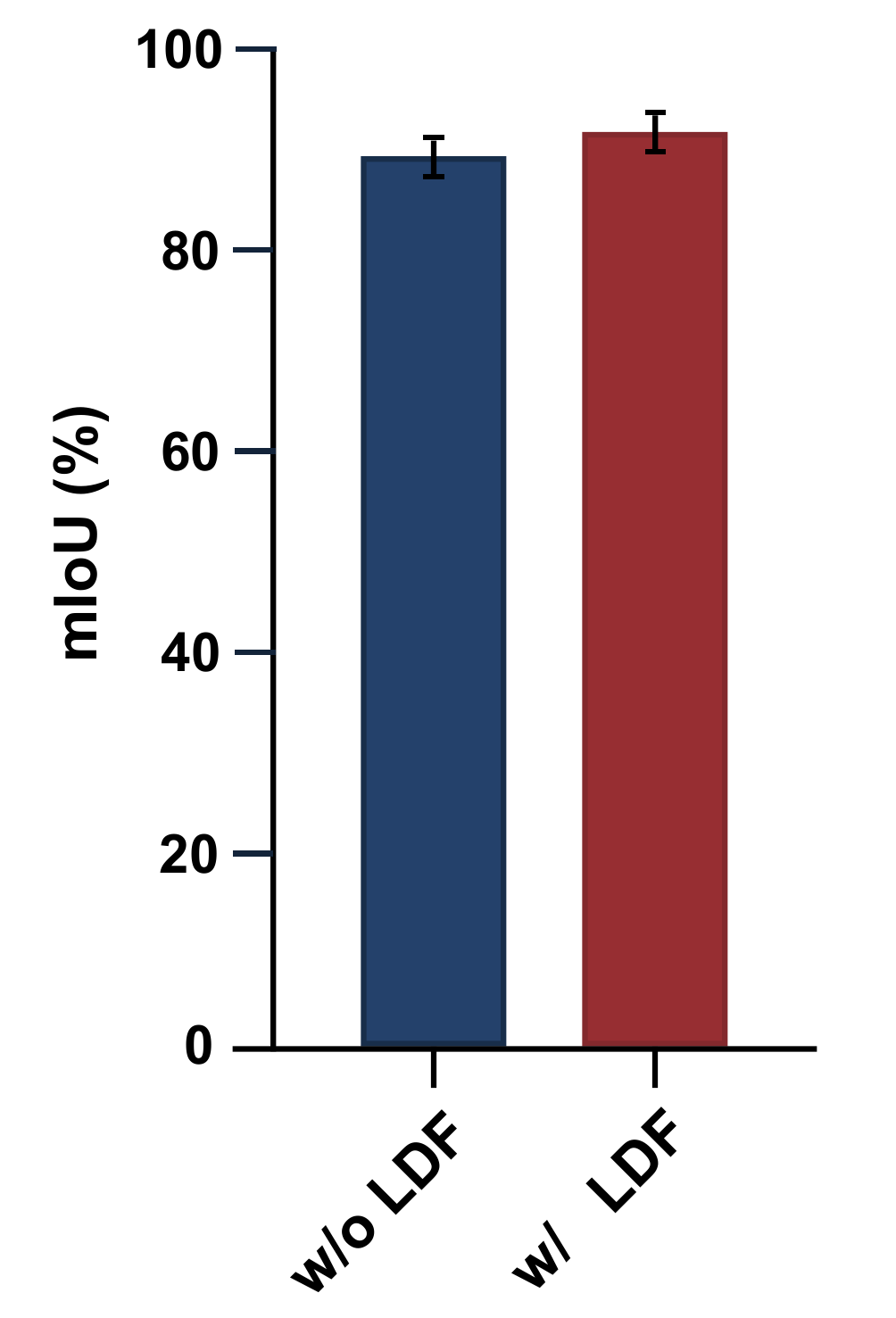}
	}\hspace{0.02\linewidth}
	\subfigure[]{
		\includegraphics[width=0.14 \linewidth]{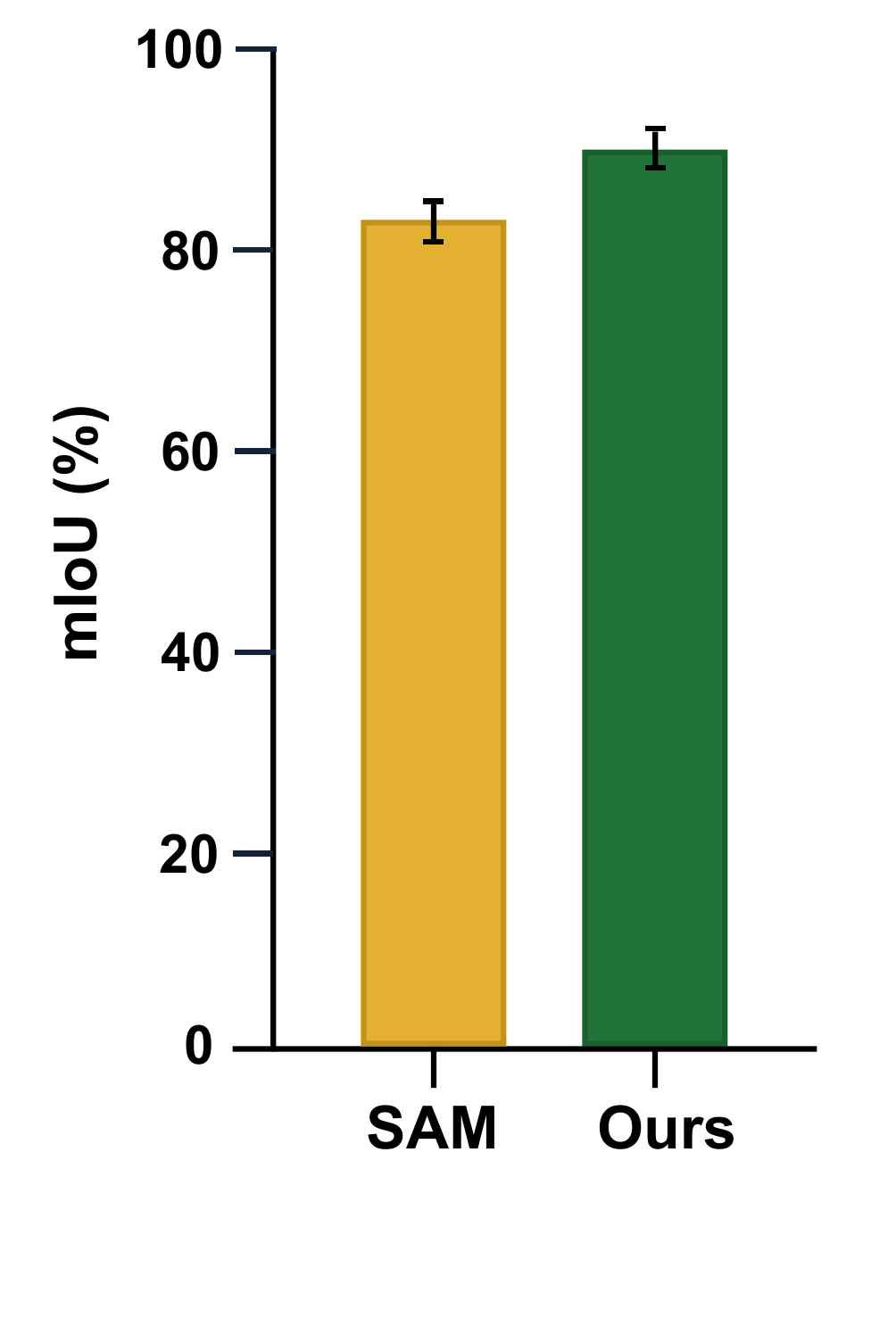}
	}\hspace{0.02\linewidth}
	\subfigure[]{
		\includegraphics[width=0.27 \linewidth]{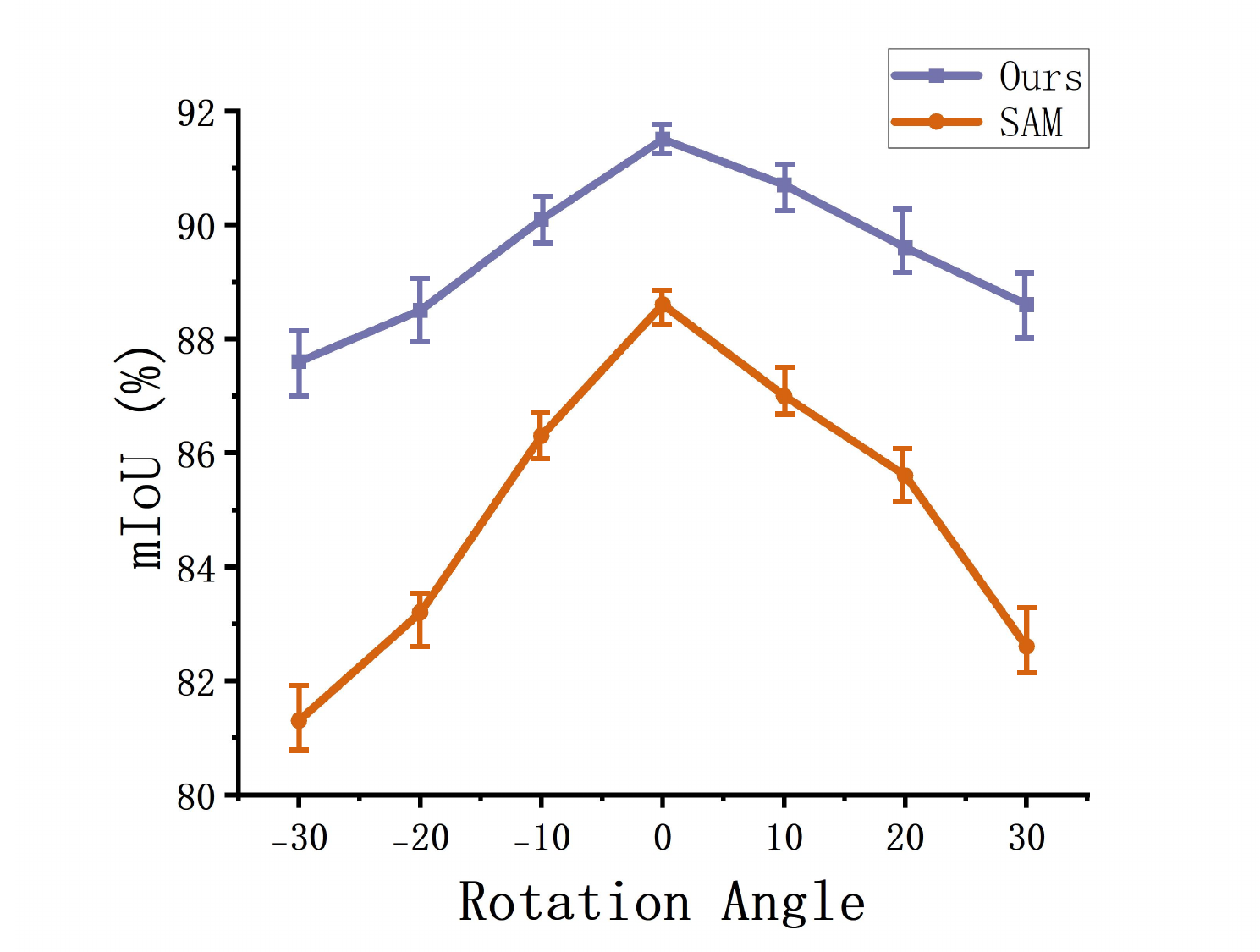}
	}
	\caption{In (a), we compare the FPS performance of our method with SAM, HQ-SAM, and EfficientSAM across different input image sizes. In (b), we present a quantitative evaluation of the impact of low-level detailed feature (LDF) aggregation on the model's mIoU. In (c), we analyze the comparative mIoU performance of our method and SAM under Gaussian noise conditions (standard deviation of 25). In (d), we illustrate the trend in mIoU variations between our method and SAM under random rotation conditions (angle range from -30° to 30°).}
	\label{figGroup}
\end{figure*}

\textbf{Bidirectional SSM.} 
In the third stage, we replaced the standard module with a Mamba block to assess the performance of the bidirectional sequence block in dental image segmentation tasks.
We conducted a comparative analysis of different strategies, as summarized in Table \ref{tab:bidirectional_comparison}. Specifically, when a standard Mamba block was directly applied, the model achieved an mIoU of 89.1\% on the dental segmentation dataset. By incorporating the bidirectional SSM module and 1D convolution separately, the mIoU improved to 90.7\% and 90.9\%, respectively. This improvement suggests that the bidirectional SSM and 1D convolution provide complementary benefits.
The bidirectional SSM enhances global contextual aggregation from both scanning directions, while the 1D convolution helps preserve local continuity along the sequence. Based on these results, we adopt the combination of the bidirectional SSM layer and 1D convolution as the final configuration for the Mamba block in the third stage, as it yields the best empirical performance among the evaluated variants.

\begin{table}[htbp]
\centering
\begin{tabular}{lcc}
\toprule
Variant & mIoU & mBIoU \\
\midrule
No Gate & 90.8 $\pm$ 0.3 & 87.2 $\pm$ 0.3 \\
Shared Gate & 91.4 $\pm$ 0.2 & 87.9 $\pm$ 0.3 \\
Dual Gate (Ours) & \cellcolor{blue!10}\textbf{91.9 $\pm$ 0.2} & \cellcolor{blue!10}\textbf{88.7 $\pm$ 0.3} \\
\bottomrule
\end{tabular}
\caption{\textbf{Ablation study on the gating mechanism. Results are reported as mean $\pm$ standard deviation across 5 folds.}}
\label{tab:ablation_gate}
\end{table}

\textbf{The Gating Mechanism.}
To evaluate the effect of the gating mechanism, we conducted ablation experiments on the DSD. 
Three variants were compared: 
(1) \textbf{No Gate}, where forward and backward features are directly summed without gating; 
(2) \textbf{Shared Gate}, where a single gating signal is applied to both branches; 
(3) \textbf{Dual Gate} (ours), where two independent gates modulate the forward and backward branches separately. 
As shown in Table~\ref{tab:ablation_gate}, removing the gates leads to a notable drop in both mIoU and mBIoU, while using a shared gate partially alleviates the degradation. The dual-gate design achieves the best performance, suggesting that independent gating of forward and backward branches helps retain salient dependencies while suppressing redundant responses.

\textbf{Efficiency Analysis.}
As shown in Fig.~\ref{figGroup}(a), our approach achieves higher efficiency than the compared methods, which can be attributed to the optimized backbone design. FPS comparisons with SAM, HQ-SAM, and EfficientSAM, all of which adopt Transformer-based architectures, indicate that their latency increases quadratically with input resolution. In contrast, our Mamba-based design exhibits approximately linear latency growth with respect to input size, making it more scalable for high-resolution inputs.

\begin{table}[!htbp]
\centering
\begin{tabular}{cccc}
\toprule
Approach & FPS$\uparrow$  & Memory(MB)$\downarrow$   & Flops(G)$\downarrow$   \\
\midrule
MedSAM~\citep{Ma_2024} & 28.5 & 3100 & 34.2 \\
Adapter-SAM~\citep{wu2025medical} & 31.5 & 3050 & 33.2 \\
SegFormer~\citep{segformer} & 43.1 & 2620 & 22.6 \\
Swin-Unet~\citep{cao2022swin} & 22.7 & 3450 & 47.8 \\
\textbf{Ours} & \cellcolor{blue!10}\textbf{52.3} & \cellcolor{blue!10}\textbf{1860} & \cellcolor{blue!10}\textbf{12.5} \\
\bottomrule
\end{tabular}
\caption{\textbf{Efficiency comparison of different methods in terms of FPS, GPU memory usage, and FLOPs on the DSD.}}
\label{Qcom}
\normalsize
\end{table}

Although some components, like the fusion decoder and bidirectional modeling, are not strictly linear, the linear-time BSB greatly reduces computation. As Table~\ref{Qcom} shows, our model achieves the best balance of speed, memory, and complexity, running at 52.3 FPS, faster than Adapter-SAM (31.5 FPS), MedSAM (28.5 FPS), Swin-Unet (22.7 FPS), and SegFormer, while using less GPU memory, highlighting its potential for real-time, resource-limited clinical applications.

\begin{table}[h]
\centering
\resizebox{\linewidth}{!}{
\begin{tabular}{lccc}
\hline
Module & FLOPs (G) & Percentage of Total & Theoretical Complexity \\
\hline
Encoder (Stage 1 \& 2)      & 2.1  & 16.8\% & $O(N)$ \\
Encoder (Stage 3 -- BSB)    & 7.8  & 62.4\% & $O(N)$ \\
Decoder (Fusion \& Upsampling) & 1.9  & 15.2\% & $O(N\log N)$ \\
Decoder (Mask Prediction)   & 0.7  & 5.6\%  & $O(N^{3/2})$ \\
\hline
\textbf{Total}              &\cellcolor{blue!10}\textbf{12.5} & \cellcolor{blue!10}\textbf{100}\%  & \cellcolor{blue!10}\textbf{$\sim O(N)$ }\\
\hline
\end{tabular}}
\caption{\textbf{Theoretical FLOPs breakdown of different modules (input resolution $640\times480$).}}
\label{tab:complexity_breakdown}
\end{table}

To analyze computational complexity, we break down FLOPs at 640×480 resolution (0.31M pixels). As Table \ref{tab:complexity_breakdown} shows, the encoder's third-stage Bidirectional Sequence Block (BSB) dominates, contributing nearly two-thirds of total FLOPs. Although decoder modules have higher theoretical complexity, their absolute FLOPs remain lower, meaning model scaling is mainly determined by the linear-time BSB stage.

\begin{table}[h]
\centering
\begin{tabular}{cccc}
\toprule
Resolution & Pixels (M) & FLOPs (G) & Runtime (ms) \\
\midrule
320×240   & 0.08 & 3.2   & 2.1   \\
640×480   & 0.31 & 12.5  & 19.1  \\
1024×768  & 0.79 & 31.6  & 49.8  \\
1536×1152 & 1.77 & 78.2  & 126.4 \\
\bottomrule
\end{tabular}
\caption{\textbf{Empirical scaling of FLOPs and runtime for our method at different input resolutions.}}
\label{tab:scaling}
\end{table}

In addition to FPS, memory, and FLOPs, we analyzed the scaling behavior of our method. 
The bidirectional SSM stage is near-linear with respect to token length, while decoder, fusion, and upsampling modules introduce mild overhead. 
As shown in Table~\ref{tab:scaling}, FLOPs grow from 3.2G at $320\times240$ to 78.2G at $1536\times1152$, with runtime increasing from 2.1\,ms to 126.4\,ms. 
Log–log fitting gives slopes of 1.15 (FLOPs) and 1.25 (runtime), indicating that the method is theoretically near-linear but practically mildly super-linear due to decoder and memory costs.

\begin{figure}[htbp]
	\centering 
	\begin{minipage}{\linewidth}
		\centering
		\includegraphics[width=0.85\linewidth]{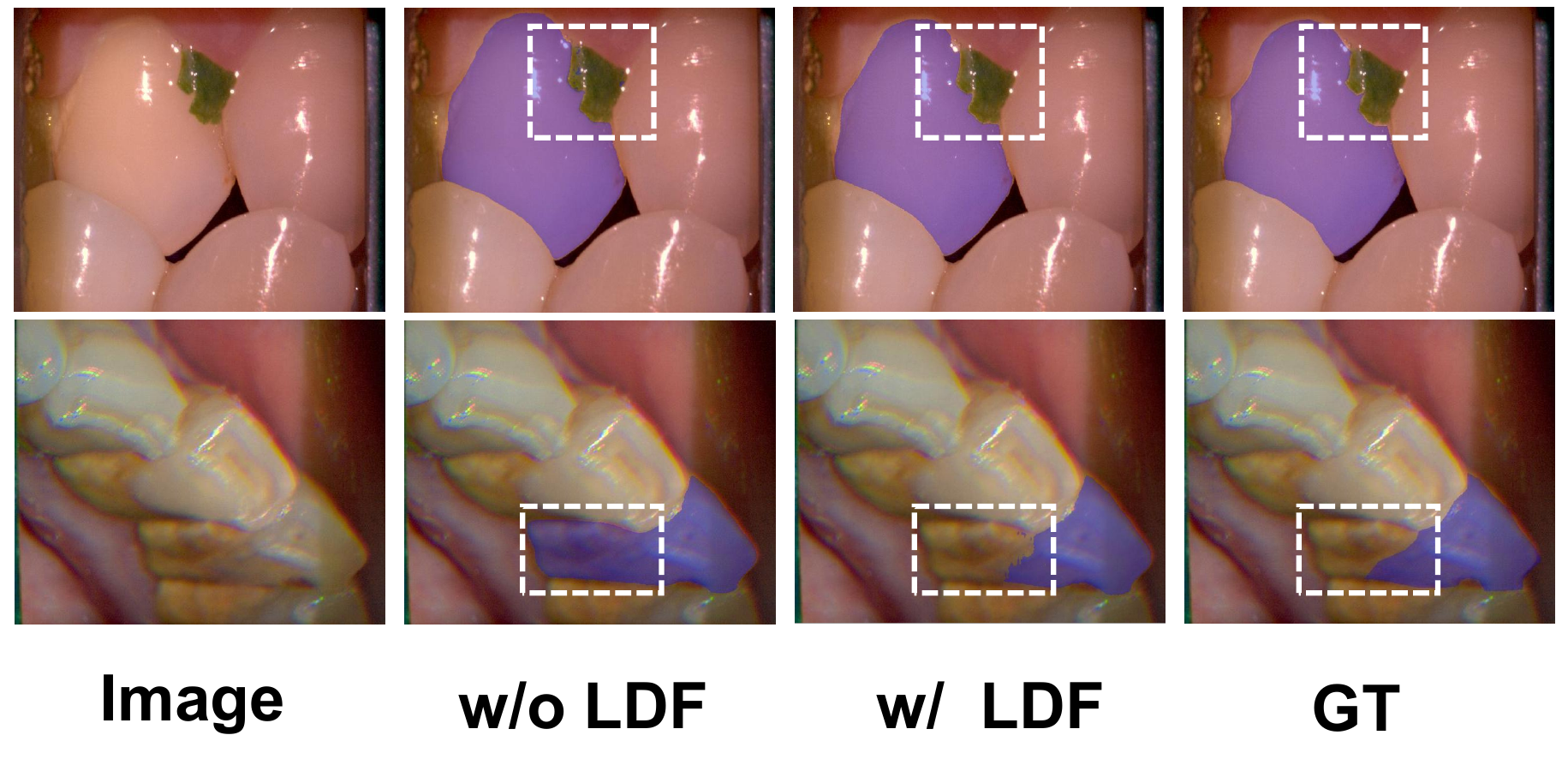} % 左图
	\end{minipage}
	\caption{\textbf{Comparison of aggregation of low-level detailed features (LDF).} The results show that aggregating low-level detail features effectively mitigates noise from artifacts such as dental calculus and food residues, producing segmentation masks that are closer to the ground truth (GT). The blue regions represent the generated masks after segmentation, and the white dashed boxes highlight the differences in the segmentation results. }
	\label{figCX}
\end{figure}

\textbf{Aggregation of low-level detailed feature (LDF).} In computer vision tasks, particularly in dental image processing, the complexity of oral environments and the similarity between tissues necessitate aggregating low-level features to extract rich spatial information, thereby improving segmentation performance. To evaluate the contribution of low-level features to segmentation accuracy, we designed an ablation experiment comparing the integration and exclusion of low-level features in the mask decoder. Fig.~\ref{figCX} illustrates the visual results of the experiment, clearly demonstrating that even in the presence of noise such as dental calculus and food debris in dental images, the aggregation of low-level features can still significantly enhance segmentation accuracy. Furthermore, the quantitative results presented in Fig.~\ref{figGroup}(b) indicate that incorporating low-level detailed features increases the mIoU by 2.7\% compared to not incorporating them. This improvement can be attributed to the fact that low-level features preserve fine-grained spatial cues and boundary information, which are often lost when relying solely on high-level semantic representations. In contrast, excluding these features leads to insufficient detail recovery, making the model more susceptible to noise and tissue ambiguity.

\begin{figure}[htbp]
	\centering 
	\begin{minipage}{\linewidth}
		\centering
		\includegraphics[width=0.8\linewidth]{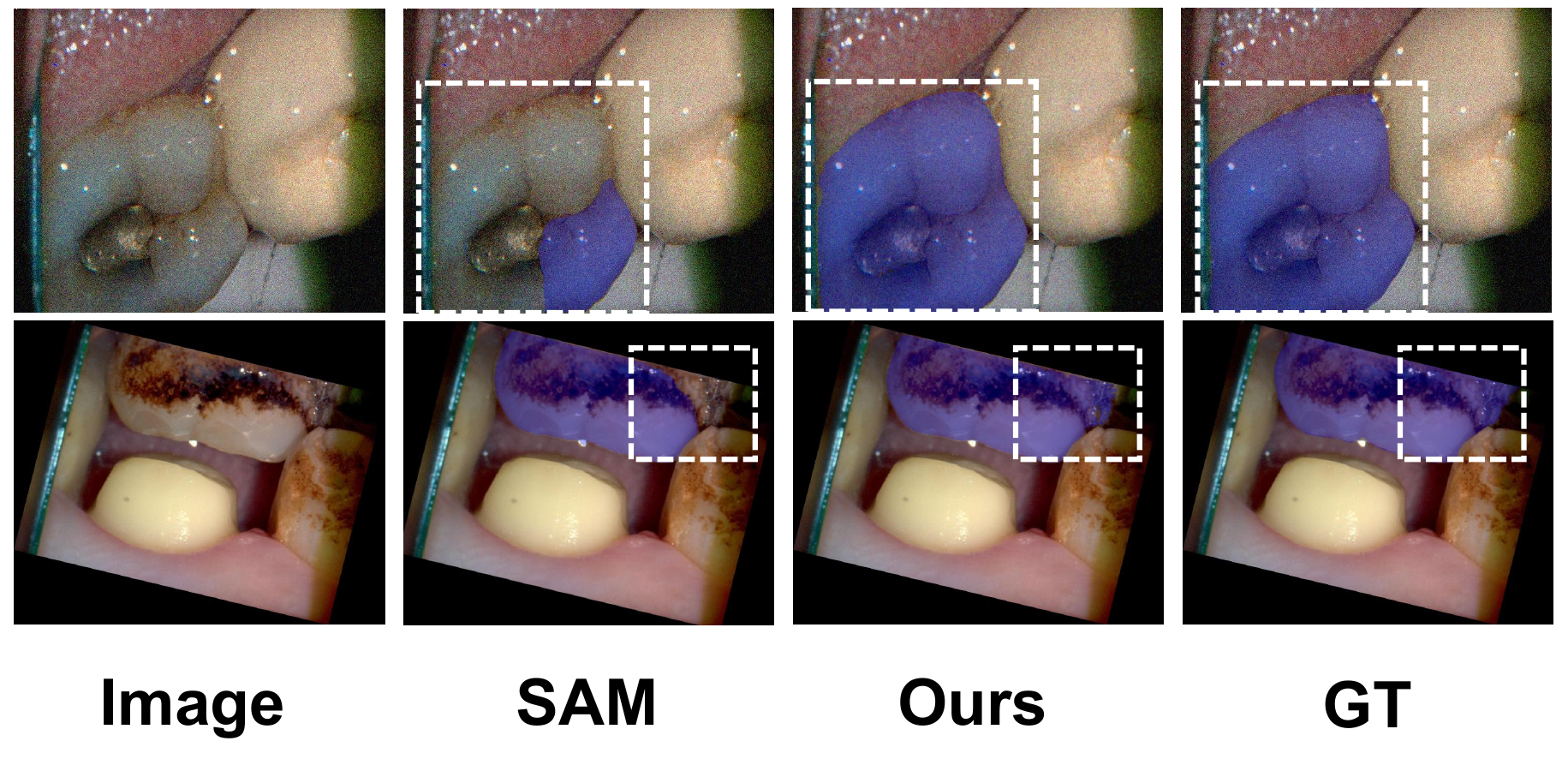} % 左图
	\end{minipage}
	\caption{\textbf{We conducted a comparative analysis with SAM under conditions of Gaussian noise and random rotation.} The results indicate that on dental images subjected to Gaussian noise and random rotation, the segmentation masks generated by our method are more complete and refined. The blue regions represent the generated masks after segmentation, and the white dashed boxes highlight the differences in the segmentation results.}
	\label{figRobust}
\end{figure}

\textbf{Robustness Analysis.} To evaluate robustness in complex environments, we adopted SAM as a reference model and applied two types of perturbations, namely Gaussian noise and random rotation. Gaussian noise was utilized to simulate pixel-level corruption commonly encountered in real-world scenarios, with zero mean and a standard deviation of 25 (in pixel intensity units, range [0,255]). Random rotation was applied to assess the impact of varying camera angles on model segmentation performance, with rotation angles ranging from -30° to +30°. As shown in Fig.~\ref{figRobust}, comparative analysis with SAM demonstrates that under conditions of Gaussian noise and random rotation, the proposed method achieves more precise and detailed segmentation results. Specifically, as depicted in Fig.~\ref{figGroup}(c), the mIoU of the proposed method outperforms SAM by 6.2\% under Gaussian noise conditions, demonstrating strong noise robustness. Furthermore, Fig.~\ref{figGroup}(d) illustrates that the proposed method exhibits high robustness across different rotation angles under random rotation conditions, with a more stable performance curve and superior overall results.

\subsection{Evaluation of Tooth Segmentation}
We evaluated various image segmentation methods on the DSD and the OralVision dataset, maintaining consistent settings for a fair comparison. 
These methods\citep{Ke_2023,xiong2024efficientsam,ren2024grounded,Ma_2024,wu2025medical} are based on the SAM model and utilize Transformer architecture and attention mechanisms to process image data. 
They can produce segmentation masks for target objects based on user-provided prompts, including clicks, bounding boxes, or textual inputs.
By introducing new visual perception techniques, these methods \citep{zhu2024vision,duan2024vision,ma2024u} enhance the understanding and analysis of image data while maintaining computational efficiency. Our method shows better results on the DSD than the aforementioned methods in Table \ref{tabSegmentation}. 

\begin{table}[htbp]
	\centering
	\footnotesize
	\begin{tabular}{lcccc}
		\toprule
		\multirow{2}{*}{Approach} & \multicolumn{2}{c}{DSD} & \multicolumn{2}{c}{OralVision} \\
		\cmidrule(lr){2-3} \cmidrule(lr){4-5}
		& mIoU & mBIoU & mIoU & mBIoU \\
		\midrule
		SAM~\citep{Kirillov_2023}   & 88.7 ± 0.4 & 85.3 ± 0.5 & 88.1 ± 0.3 & 85.0 ± 0.4 \\
		HQ-SAM~\citep{Ke_2023}      & 91.2 ± 0.3 & 88.1 ± 0.4 & 90.3 ± 0.3 & 87.2 ± 0.3 \\
		EfficientSAM~\citep{xiong2024efficientsam} & 87.2 ± 0.5 & 84.4 ± 0.5 & 86.9 ± 0.4 & 83.8 ± 0.3 \\
		GroundedSAM~\citep{ren2024grounded} & 88.6 ± 0.4 & 85.5 ± 0.3 & 87.6 ± 0.3 & 84.7 ± 0.3 \\
		MedSAM~\citep{Ma_2024}  & 89.0 ± 0.3 & 86.0 ± 0.4 & 88.5 ± 0.3 & 85.8 ± 0.3 \\
		Adapter-SAM~\citep{wu2025medical} & 88.5 ± 0.4 & 85.5 ± 0.4 & 88.0 ± 0.3 & 85.0 ± 0.4 \\
        \midrule
        SegFormer~\citep{segformer} & 89.8 ± 0.3 & 86.3 ± 0.3 & 89.1 ± 0.4 & 85.8 ± 0.3 \\
        Swin-Unet~\citep{cao2022swin} & 89.3 ± 0.3 & 86.0 ± 0.3 & 88.9 ± 0.4 & 85.5 ± 0.4 \\
		\midrule
		Vim~\citep{zhu2024vision} & 90.6 ± 0.3 & 86.4 ± 0.4 & 89.9 ± 0.4 & 86.3 ± 0.3 \\
		VRWKV~\citep{duan2024vision} & 89.8 ± 0.4 & 85.7 ± 0.3 & 89.2 ± 0.3 & 84.8 ± 0.3 \\
		U-Mamba~\citep{ma2024u} & 90.1 ± 0.3 & 85.8 ± 0.4 & 90.0 ± 0.3 & 85.2 ± 0.3 \\
		\midrule
		\textbf{Ours} & \cellcolor{blue!10}\textbf{91.9 ± 0.2} & \cellcolor{blue!10}\textbf{88.7 ± 0.3} & \cellcolor{blue!10}\textbf{91.4 ± 0.3} & \cellcolor{blue!10}\textbf{88.1 ± 0.2} \\
		\bottomrule
	\end{tabular}
	\caption{\textbf{Experimental results on the test sets of the DSD and OralVision datasets.} All values are reported as mean ± standard deviation.}
	\label{tabSegmentation}
\end{table}

In comparison to HQ-SAM, our approach yields a 0.7\% improvement in mIoU on the DSD and a 1.1\% improvement on the OralVision dataset, demonstrating its effectiveness in optimizing dental image segmentation results. Additionally, compared to methods using linear mechanisms such as U-Mamba and VRWKV, our approach significantly outperforms them in both mIoU and mBIoU on the two dental datasets. 

\begin{table}[!htbp]
\centering
\begin{tabular}{lccc}
\toprule
Class & SAM & HQ-SAM & Ours \\
\midrule
Teeth   & 90.2 $\pm$ 0.3 & 92.5 $\pm$ 0.2 & \cellcolor{blue!10} \textbf{93.6 $\pm$ 0.3 }\\
Gums    & 86.3 $\pm$ 0.2 & 88.7 $\pm$ 0.3 & \cellcolor{blue!10} \textbf{89.8 $\pm$ 0.2} \\
Lips    & 87.4 $\pm$ 0.3 & 89.9 $\pm$ 0.2 & \cellcolor{blue!10} \textbf{91.0 $\pm$ 0.2 }\\
Cheeks  & 87.0 $\pm$ 0.2 & 89.5 $\pm$ 0.3 & \cellcolor{blue!10} \textbf{90.6 $\pm$ 0.3} \\
Tongue  & 89.5 $\pm$ 0.2 & 91.0 $\pm$ 0.3 & \cellcolor{blue!10} \textbf{92.0 $\pm$ 0.2}\\
\midrule
mIoU (overall) & 88.1 $\pm$ 0.3 & 90.3 $\pm$ 0.3 & \cellcolor{blue!10} \textbf{91.4 $\pm$ 0.3} \\
\bottomrule
\end{tabular}
\caption{\textbf{Per-class IoU on OralVision.}}
\label{tab:oral_per_class}
\end{table}

As shown in Fig.~\ref{figGroup}(a), our method outperforms vision transformer–based approaches (e.g., SAM, HQ-SAM, EfficientSAM) in efficiency and latency, with its FPS advantage growing at higher resolutions. Although Mamba was designed for long-sequence NLP, it remains effective for moderate-resolution medical segmentation. Table~\ref{Qcom} and Table~\ref{tabSegmentation} show that our approach achieves competitive accuracy while offering faster inference speed and lower memory usage among the compared methods, meeting the dual demands of precision and real-time responsiveness. It should be emphasized that prompt usage was carefully controlled for fairness. SAM-based methods, including our approach, were all tested with identical bounding box prompts, whereas non-SAM baselines followed their native inference protocols without prompts. This protocol ensures that the performance differences shown in Table~\ref{tabSegmentation} truly reflect architectural strengths rather than prompt bias. Table~\ref{tab:oral_per_class} reports per-class IoU (mean ± standard deviation across 5 folds) for the OralVision test set. The improvements across all classes indicate that hierarchical feature representation and bidirectional context modeling help preserve fine-grained boundaries and inter-class distinctions, whereas single-scale or unidirectional designs are more prone to losing local details or global consistency, resulting in inferior per-class performance.

\begin{figure}[htbp]
	\centering
	\includegraphics[width=0.75 \linewidth]{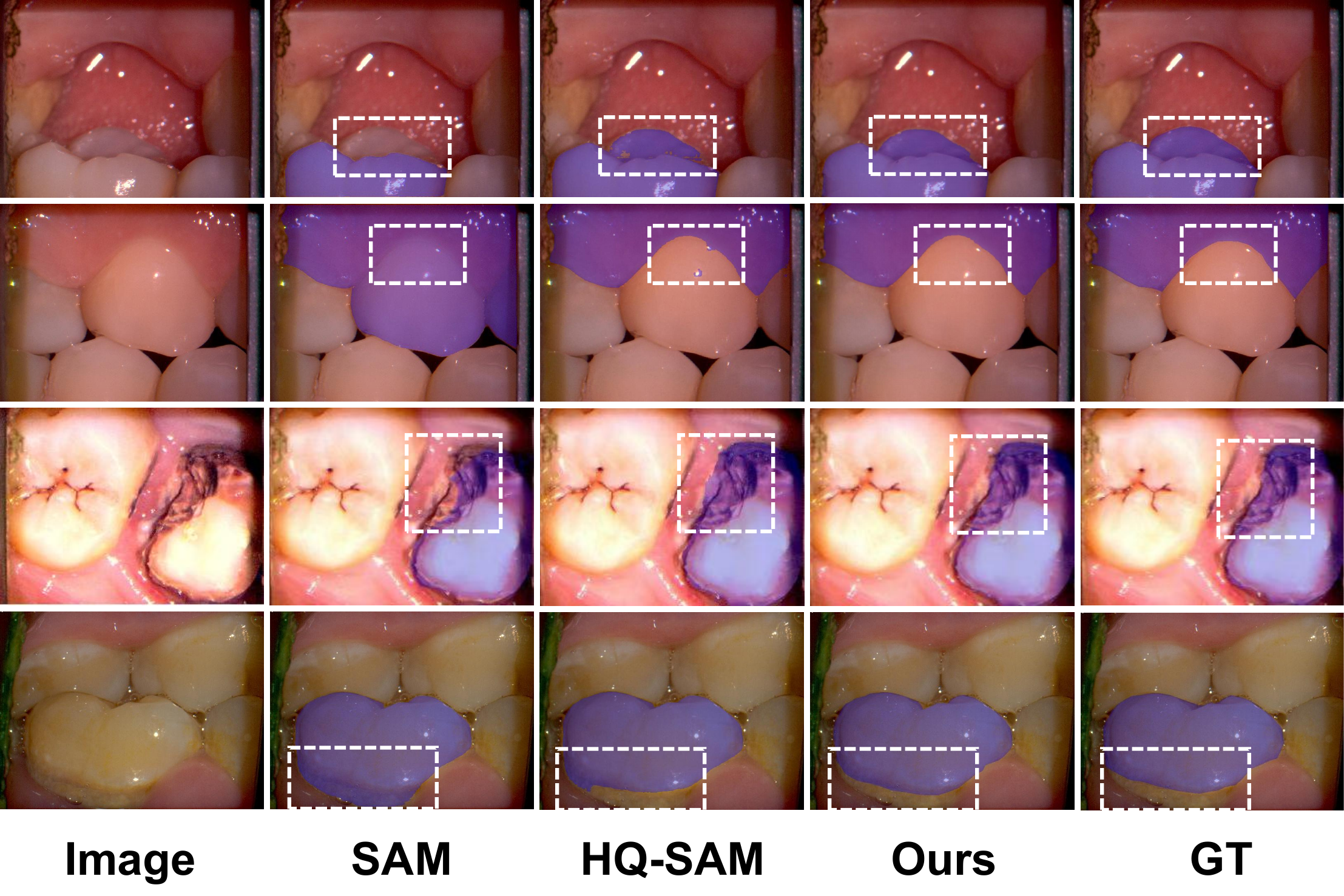}
	\caption{\textbf{The visualization results of SAM, HQ-SAM, and our method on the DSD.} Compared to other methods, our approach generates more fine-grained segmentation masks. The blue area represents the mask generated after image segmentation. The white dashed box highlights the differences between the segmentation masks generated by our method and those produced by other methods.}
	\label{figCompare}
	\vspace{-2mm}
\end{figure}

In comparison to SAM, HQ-SAM, and our method, the segmentation results on dental images are visualized in Fig.~\ref{figCompare}. Our approach shows more refined segmentation performance, even in the presence of complex oral environments. Noise factors such as saliva, food debris, and tartar can interfere with the segmentation process, yet our model handles these details more effectively. In the first example, for instance, the molar of the patient has multiple 3D surfaces, and some folding appears in the image. SAM mistakenly segmented the molar into two parts, while HQ-SAM failed to segment the molar completely. In dental image acquisition within oral environments, the quality of segmentation is inevitably affected by noise such as tartar reflections. In the fourth example, our method effectively mitigates the impact of such noise, producing finer-grained and higher-quality segmentation masks.

\subsection{Discussion}
% what we do
% Dental image segmentation is a critical component of digitalization in oral medicine. We introduce an effective segmentation technique rooted in Mamba, capable of producing high-quality segmentation masks by utilizing point or box prompts. A range of evaluation experiments were conducted to analyze the behavior of the proposed method on dental datasets.

In dental image segmentation, strong noise, complex tissue appearance, and the joint demand for accuracy and efficiency pose significant challenges. Experimental results show that integrating hierarchical feature modeling with bidirectional sequence-based context aggregation enables a more effective balance between segmentation precision, robustness, and computational cost. The bidirectional design improves global context consistency under noisy conditions, which is critical for stable fine-grained segmentation. This balance is particularly important for real-world oral medical applications, where both reliable delineation and fast inference are required.

% effect (good)
% The experimental findings on the DSD and OralVision Dataset further validate its effectiveness. Specifically, the method ensures segmentation quality by integrating multi-scale features through a feature pyramid and incorporating low-level features into the decoder. Benefiting from the application of linear mechanisms, we leverage Mamba blocks to achieve global feature awareness while enhancing the model's processing speed, thereby improving inference efficiency, which is beneficial for practical deployment in oral medical imaging scenarios.
% The experimental results on the DSD and OralVision datasets demonstrate that the proposed method consistently achieves superior segmentation performance across different dental scenarios. Compared with HQ-SAM, our approach shows noticeable improvements in mIoU on both datasets, indicating enhanced overall segmentation accuracy. Moreover, when compared with other models based on linear mechanisms, the proposed method maintains clear advantages in both region-level accuracy and boundary quality, suggesting that the performance gains are not obtained at the cost of structural precision. In addition to accuracy, the method also exhibits improved inference efficiency, enabling a better balance between segmentation quality and computational cost, which is particularly important for practical dental imaging applications.

The experimental results on the DSD and OralVision datasets demonstrate that the proposed method consistently achieves superior segmentation performance across different dental scenarios. Compared with HQ-SAM, our approach shows noticeable improvements in mIoU on both datasets, indicating enhanced overall segmentation accuracy. Moreover, compared with other models based on linear mechanisms, our method achieves better region-level accuracy and boundary quality without sacrificing structural precision. Ablation results further show that both components are independently effective: using the BSB yields an mIoU of 90.9\%, while incorporating LDF achieves an mIoU of 91.2\%, confirming their respective contributions to contextual modeling and fine-grained detail preservation. In addition, the proposed method maintains favorable inference efficiency, making it suitable for practical dental imaging applications.

% why good in deep analysis?
% In dental image segmentation, we introduced a three-stage feature pyramid structure, where the high-resolution features generated in the first two stages are utilized in the decoder to refine segmentation masks and mitigate information loss. By leveraging rich visual contextual information to capture low-level features and integrating these with the prompt encoder, the method effectively refines masks and produces high-quality segmentation results. The utilization of multi-scale feature maps enhances the model's focus on spatial information, which contributes to improved robustness under complex oral imaging conditions.

In dental image segmentation, the core advantage of the proposed architecture lies in its hierarchical representation, which effectively regulates the flow of information across different semantic levels. Low-level features preserve precise geometric cues and boundary continuity, which are particularly vulnerable to degradation in noisy oral environments, whereas high-level features provide more stable semantic representations to suppress background interference. Their complementary interaction enables the decoder to selectively recover structurally consistent target regions without amplifying noise. Moreover, the multi-scale representation enhances the model’s adaptability to variations in tooth morphology and surrounding tissues, which is crucial for handling noise, illumination changes, and tissue similarity. 

% Additionally, the introduction of a bidirectional sequence block within the linear mechanism helps alleviate the potential loss of contextual information caused by unidirectional scanning. By combining forward and backward scans, this block enhances visual context understanding and spatial awareness while retaining Mamba’s efficiency. Unlike vision transformers and CNNs that depend on heavy 2D operations, our method maintains spatial precision through shallow convolutions and gated fusion. These observations suggest that the 1D sequence modeling strategy can preserve segmentation accuracy and spatial detail while reducing computational overhead.

The effectiveness of the bidirectional sequence block stems from its ability to balance long-range dependency modeling with spatial consistency. By jointly modeling forward and backward dependencies, the bidirectional design enables more symmetric context aggregation, reducing directional bias and stabilizing feature responses across spatial regions. Combined with lightweight convolutions and gated fusion, this design preserves fine-grained spatial details while maintaining linear computational complexity, explaining why 1D sequence modeling achieves both high segmentation accuracy and efficiency in complex oral imaging scenarios.

\begin{figure}[htbp]
	\centering
	\includegraphics[width=0.55 \linewidth]{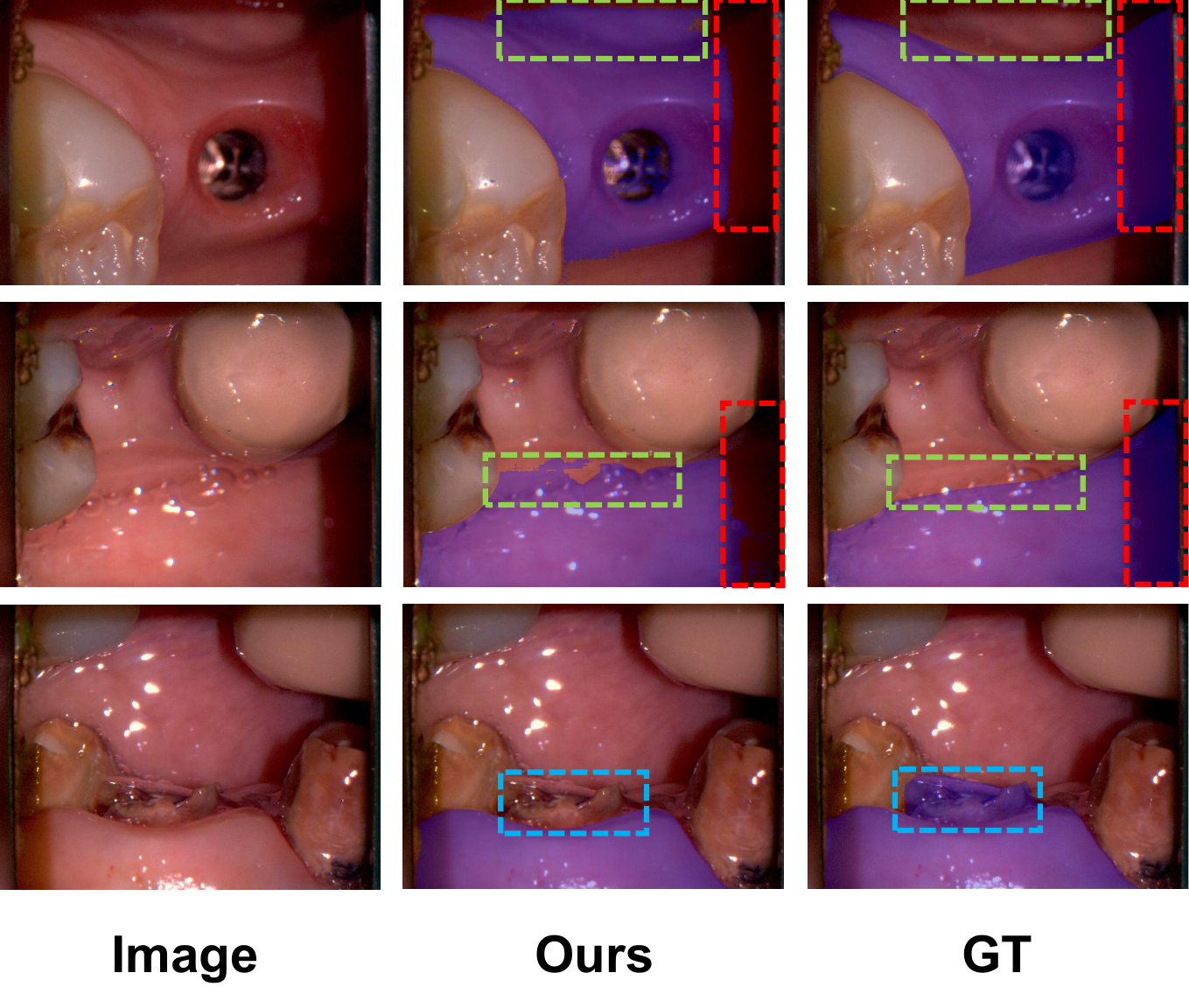}
	\caption{\textbf{Failure cases of our approach in DSD.} The blue area represents the mask generated after image segmentation. The red dashed box highlights areas where incomplete segmentation occurred due to poor image brightness. The green dashed box indicates segmentation errors caused by the similar color between the gums and cheek tissue. The blue dashed box marks regions where the model failed to segment correctly due to saliva noise and misleading folded surfaces.}
	\label{figDrawback}
\end{figure}
% drawback (figure) and reason
Although the proposed method achieves strong performance in most cases, the failure cases reveal several limitations when confronted with challenging oral imaging conditions. As shown in Fig.~\ref{figDrawback}, the segmentation results heavily depend on image brightness, and when brightness is low, the quality of the generated segmentation masks may deteriorate. Furthermore, when the colors of the gums and cheek tissues are similar, the model might be misled, leading to segmentation errors.

% future work to solve drawback
In future work, incorporating text prompts may further enhance the semantic guidance provided to the segmentation model. Enhancing the model’s performance will enable it to capture richer semantic information, thereby producing higher-quality segmentation masks. Efforts will also be made to expand the OralVision dataset by collecting more samples to increase data diversity and provide broader training support for the model. Additionally, we plan to evaluate the proposed method on public dental datasets beyond DSD and OralVision to further assess its generalizability and transferability across diverse clinical imaging scenarios.

\section{Conclusion}\label{conclusion}
This paper proposes a segmentation method for the dental domain based on bidirectional sequence modeling and hierarchical feature representation. Experimental results demonstrate that our approach outperforms other image segmentation methods in dental image segmentation. Specifically, by integrating both low-level features from each encoder layer and high-level features from subsequent layers into the decoder, our method achieves multi-scale feature extraction and fusion, thereby enhancing segmentation performance. Moreover, the structured state space model within the Mamba block effectively captures image features while significantly reducing computational complexity through parallel processing. However, despite these advantages, our approach still faces challenges in segmenting regions with poor lighting conditions and anatomically similar oral tissues. Nevertheless, the high-quality segmentation results obtained on two dental image datasets highlight its significant potential for clinical applications. Future research could focus on exploring the feasibility of multimodal semantic segmentation in the dental domain, improving segmentation performance under low-light conditions, and further optimizing the model for lightweight deployment.

\bibliographystyle{IEEEtran}
\bibliography{mybibfile1}

\end{document}